\documentclass[acmtog, authorversion, nonacm]{acmart}

\usepackage{booktabs} %

\usepackage{graphicx}
\usepackage{amsmath}
\usepackage{booktabs}
\usepackage[title]{appendix}

\usepackage[utf8]{inputenc}

\usepackage{graphicx}
\usepackage{wrapfig}

\usepackage[capitalize, sort]{cleveref}
\crefname{section}{Sec.}{Secs.}
\Crefname{section}{Section}{Sections}
\Crefname{table}{Table}{Tables}
\crefname{table}{Tab.}{Tabs.}

\usepackage[ruled]{algorithm2e} %
\usepackage{xhfill}

\SetAlFnt{\small}
\SetAlCapFnt{\small}
\SetAlCapNameFnt{\small}
\SetAlCapHSkip{0pt}

\acmJournal{FACMP}

\newcommand{\ignorethis}[1]{}

\makeatletter
\DeclareRobustCommand\onedot{\futurelet\@let@token\@onedot}
\def\@onedot{\ifx\@let@token.\else.\null\fi\xspace}

\def\eg{\emph{e.g}.} 
\def\ie{\emph{i.e}.}

\makeatother

\newcommand*{\rom}[1]{\expandafter\romannumeral #1}

\definecolor{mydarkblue}{rgb}{0,0.08,1}
\definecolor{mydarkgreen}{rgb}{0.02,0.6,0.02}
\definecolor{mydarkred}{rgb}{0.8,0.02,0.02}
\definecolor{mydarkorange}{rgb}{0.40,0.2,0.02}
\definecolor{mypurple}{RGB}{111,0,255}
\definecolor{myred}{rgb}{1.0,0.0,0.0}
\definecolor{mygold}{rgb}{0.75,0.6,0.12}
\definecolor{myblue}{rgb}{0,0.2,0.8}
\definecolor{mydarkgray}{rgb}{0.66,0.66,0.66}
\definecolor{amethyst}{rgb}{0.6, 0.4, 0.8}
\definecolor{orange}{rgb}{0.93,0.48,0.03}
\definecolor{blueviolet}{rgb}{0.54,0.16,0.88}
\definecolor{jazzberryjam}{rgb}{0.65, 0.04, 0.37}
\definecolor{darkg}{rgb}{0,0.6,0}

\newif\ifcolor
\colortrue

\newcommand{\etal}{{{et al}. }}

\newcommand{\w}{$\mathcal{W}$\xspace}

\title[StyleGAN]%
      {State-of-the-Art in the Architecture, Methods and Applications of StyleGAN}

\author{Amit H. Bermano}
\affiliation{\institution{Tel Aviv University} }
\author{Rinon Gal}
\affiliation{\institution{Tel Aviv University} }
\author{Yuval Alaluf}
\affiliation{\institution{Tel Aviv University} }
\author{Ron Mokady}
\affiliation{\institution{Tel Aviv University} }
\author{Yotam Nitzan}
\affiliation{\institution{Tel Aviv University} }
\author{Omer Tov}
\affiliation{\institution{Tel Aviv University} }
\author{Or Patashnik}
\affiliation{\institution{Tel Aviv University} }
\author{Daniel Cohen-Or}
\affiliation{\institution{Tel Aviv University} }

\begin{document}

\begin{abstract}

Generative Adversarial Networks (GANs) have established themselves as a prevalent approach to image synthesis. Of these, StyleGAN offers a fascinating case study, owing to its remarkable visual quality and an ability to support a large array of downstream tasks. 
This state-of-the-art report covers the StyleGAN architecture, and the ways it has been employed since its conception, while also analyzing its severe limitations. It aims to be of use for both newcomers, who wish to get a grasp of the field, and for more experienced readers that might benefit from seeing current research trends and existing tools laid out.

Among StyleGAN's most interesting aspects is its learned latent space. Despite being learned with no supervision, it is surprisingly well-behaved and remarkably disentangled. Combined with StyleGAN's visual quality, these properties gave rise to unparalleled editing capabilities.
However, the control offered by StyleGAN is inherently limited to the generator's learned distribution, and can only be applied to images generated by StyleGAN itself.
Seeking to bring StyleGAN's latent control to real-world scenarios, the study of GAN inversion and latent space embedding has quickly gained in popularity. Meanwhile, this same study has helped shed light on the inner workings and limitations of StyleGAN.
We map out StyleGAN's impressive story through these investigations, and discuss the details that have made StyleGAN the go-to generator.
We further elaborate on the visual priors StyleGAN constructs, and discuss their use in downstream discriminative tasks.
Looking forward, we point out StyleGAN's limitations and speculate on current trends and promising directions for future research, such as task and target specific fine-tuning.
\end{abstract} 



\maketitle

\section{Introduction}

\begin{figure}
\centering
\includegraphics[width=0.99\linewidth]{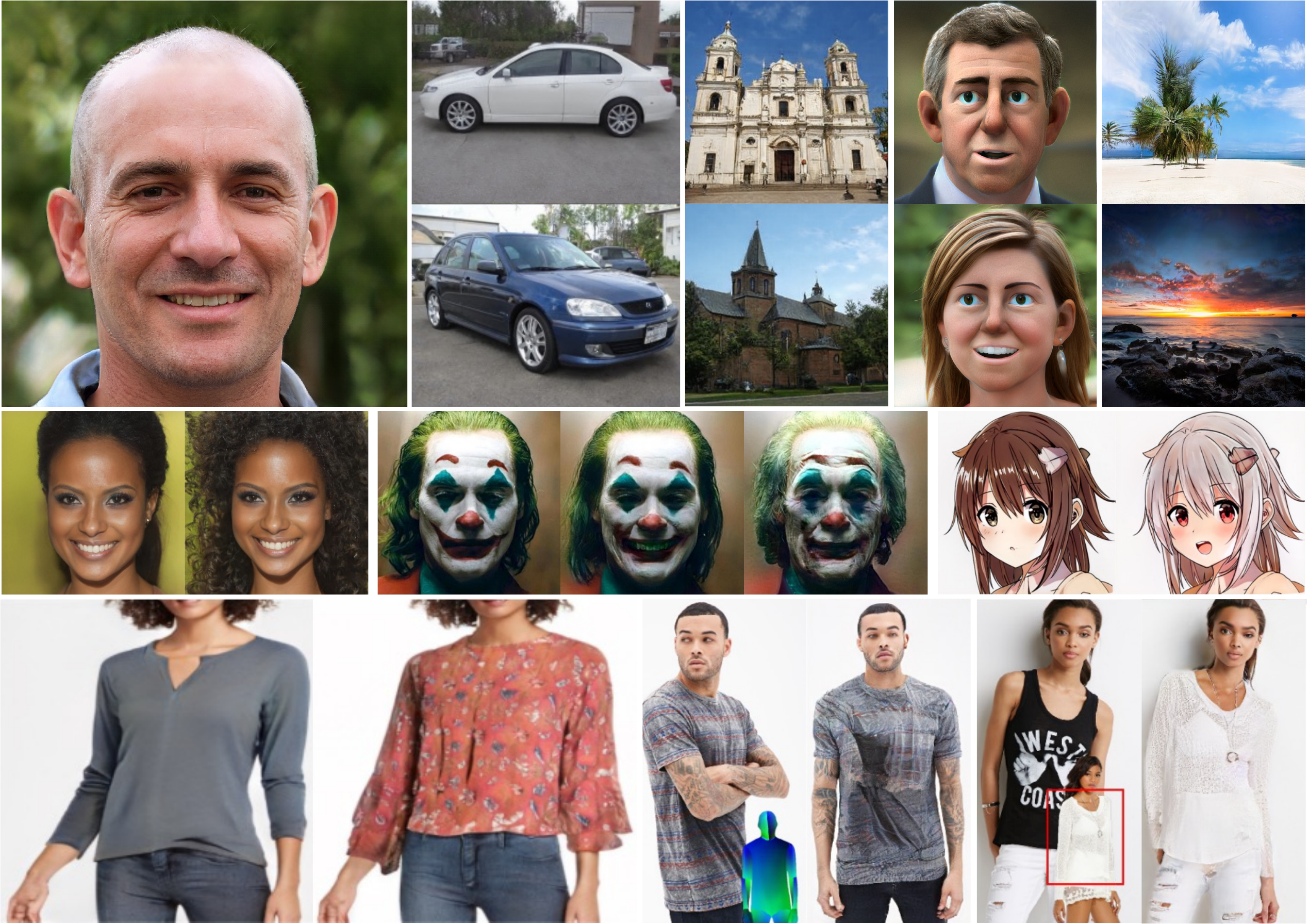}
\caption{Images synthesized by StyleGAN, its followups and derivative works.}
\vspace{-12pt}
\label{fig:teaser}
\end{figure}

The ability of GANs to generate images of phenomenal realism at high resolutions is revolutionizing the field of image synthesis and manipulation. More specifically, StyleGAN \cite{karras2019style} has reached the forefront of image synthesis, gaining recognition as the state-of-the-art generator for high-quality images and becoming the de-facto golden standard for the editing of facial images. See Figure~\ref{fig:teaser}, top for some visual examples.

\begin{figure}

\setlength{\tabcolsep}{1pt}
\centering
\begin{tabular}{ccc}
  \includegraphics[width=0.32\columnwidth]{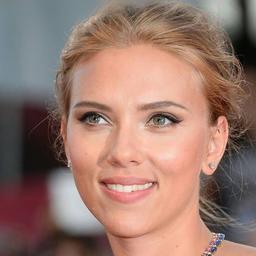} & 
  \includegraphics[width=0.32\columnwidth]{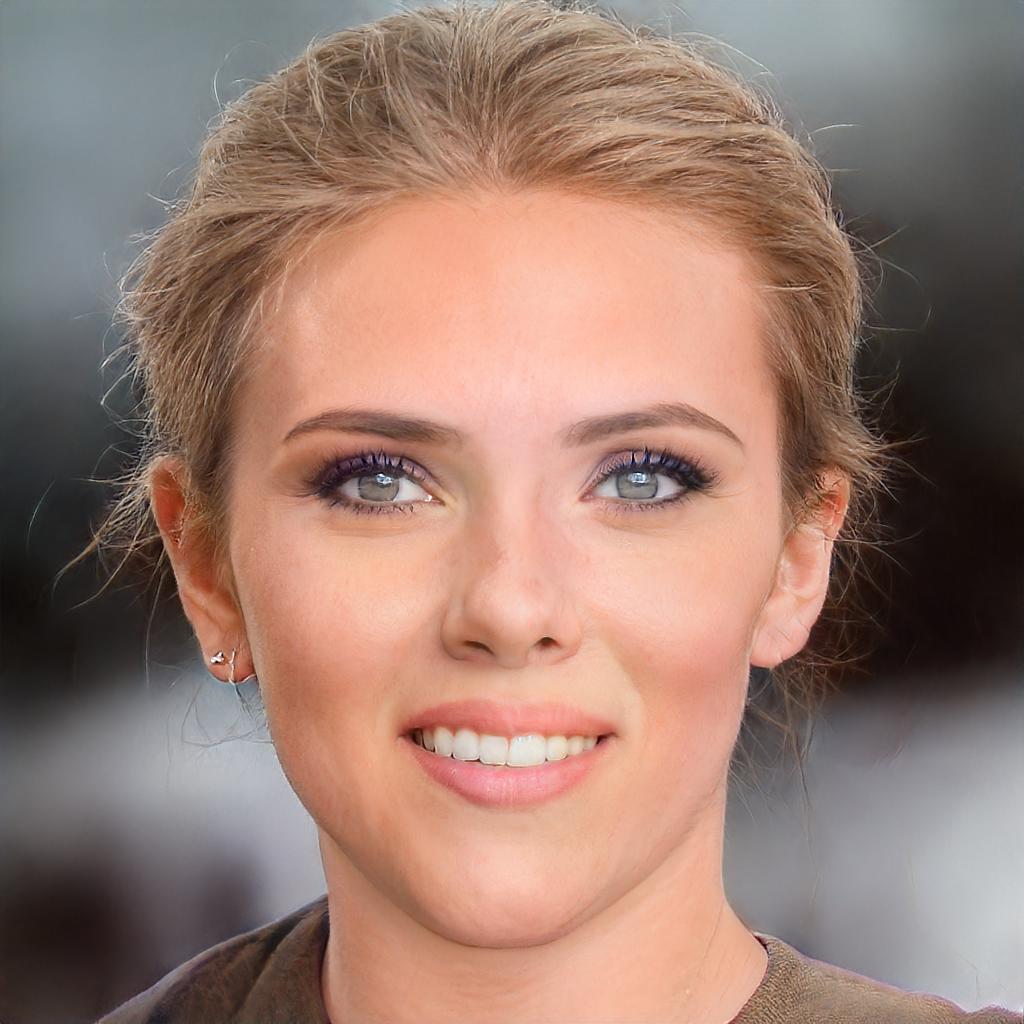} & 
  \includegraphics[width=0.32\columnwidth]{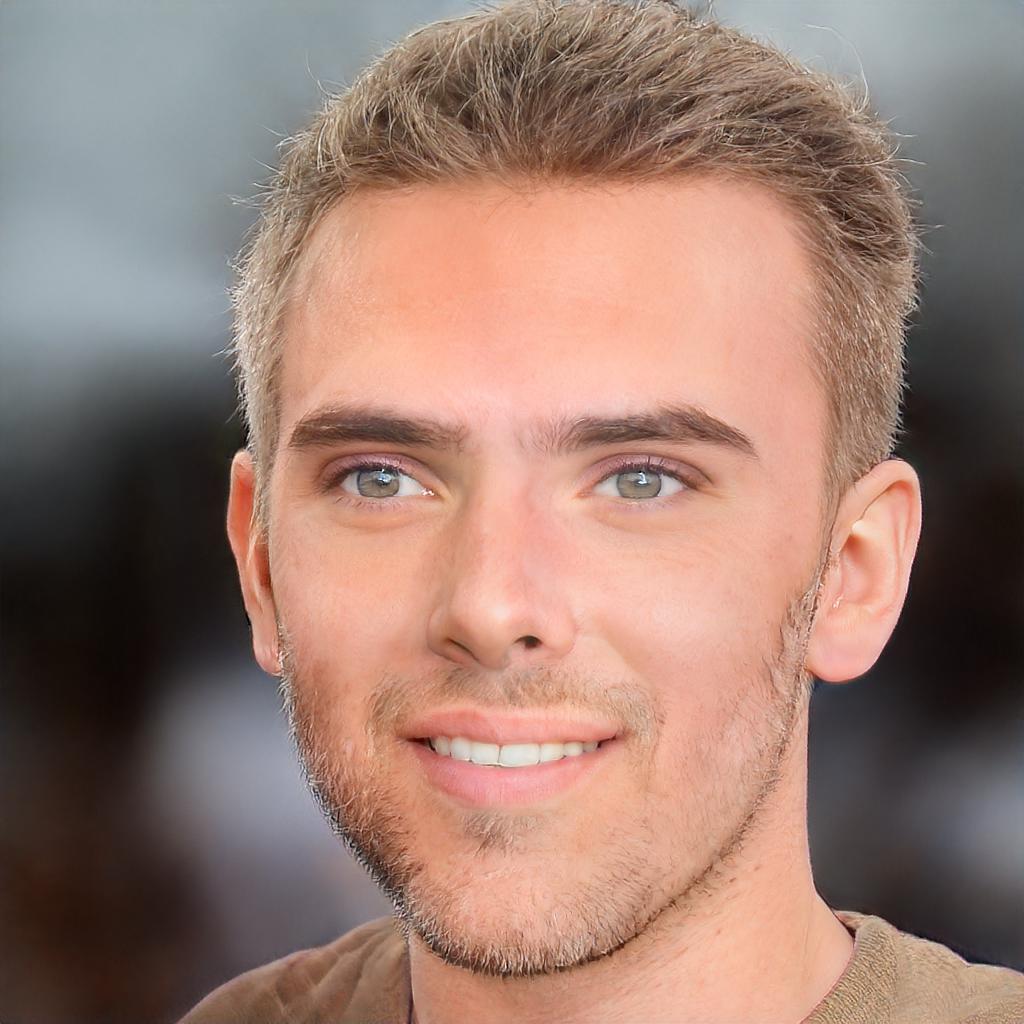} \\
  Input & Pose & Gender \\
  \includegraphics[width=0.32\columnwidth]{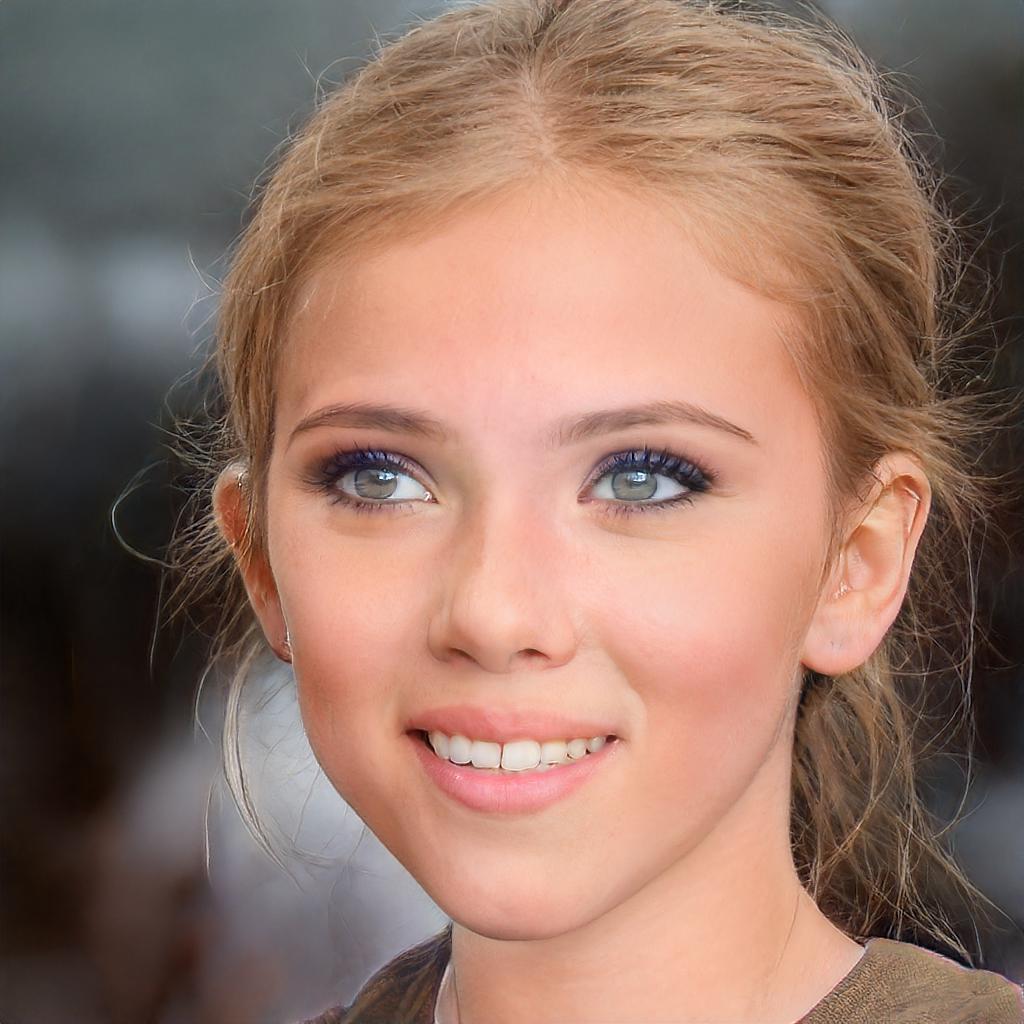} & 
  \includegraphics[width=0.32\columnwidth]{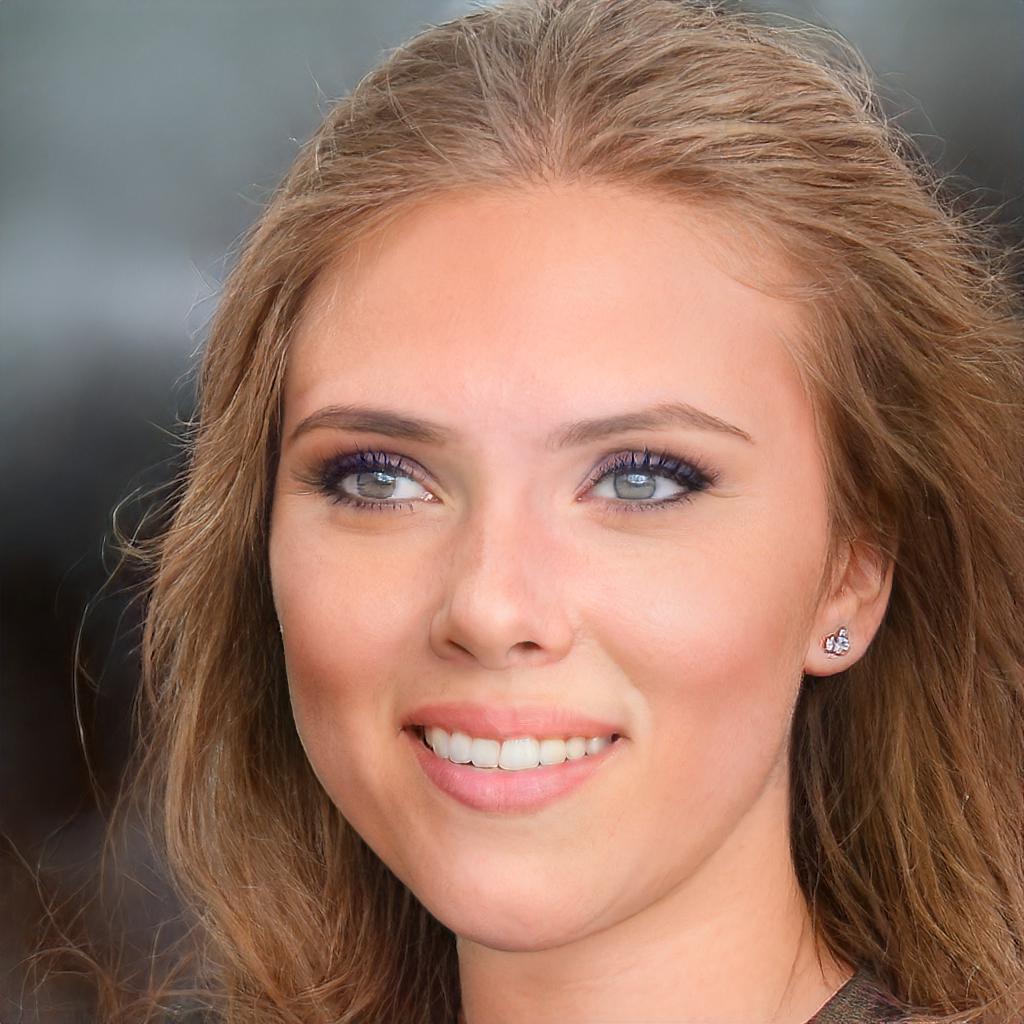} &
  \includegraphics[width=0.32\columnwidth]{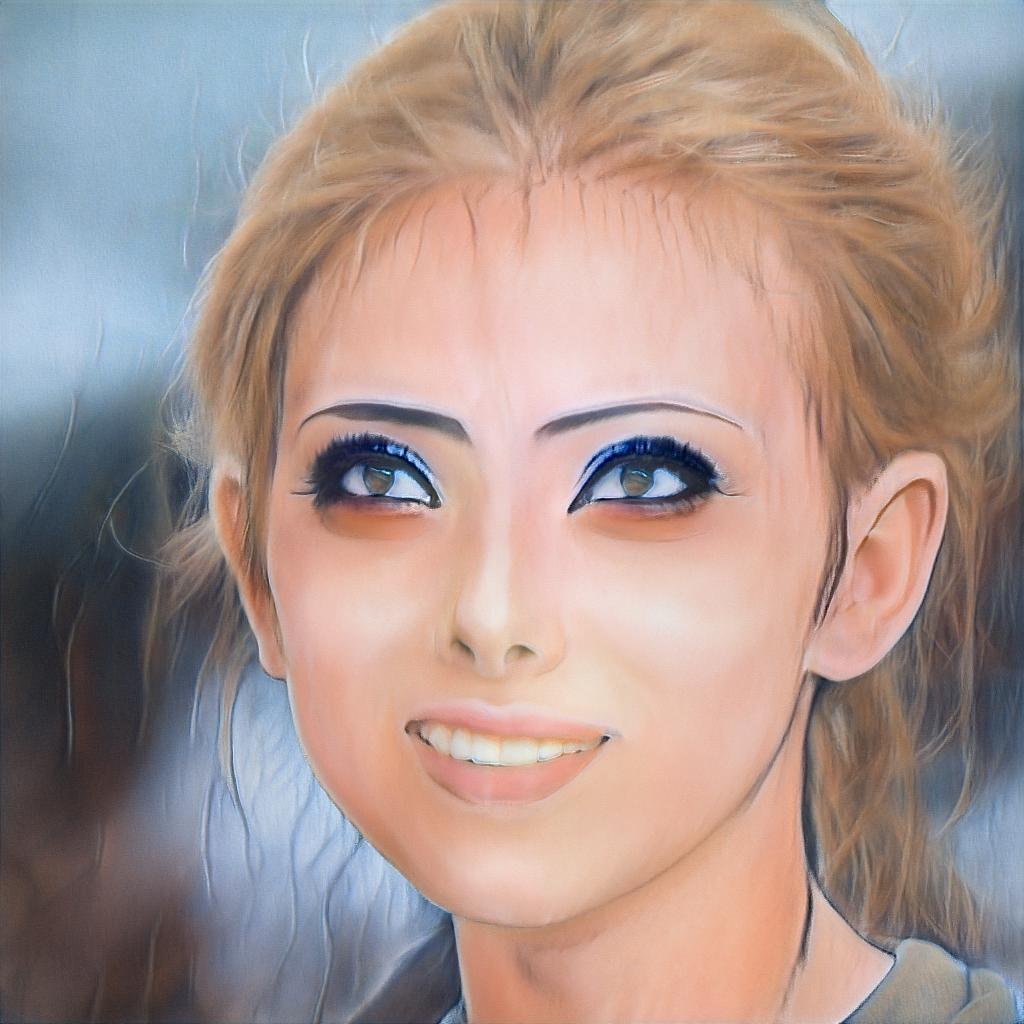}  \\
  Age & Long Hair & Anime \\
  \includegraphics[width=0.32\columnwidth]{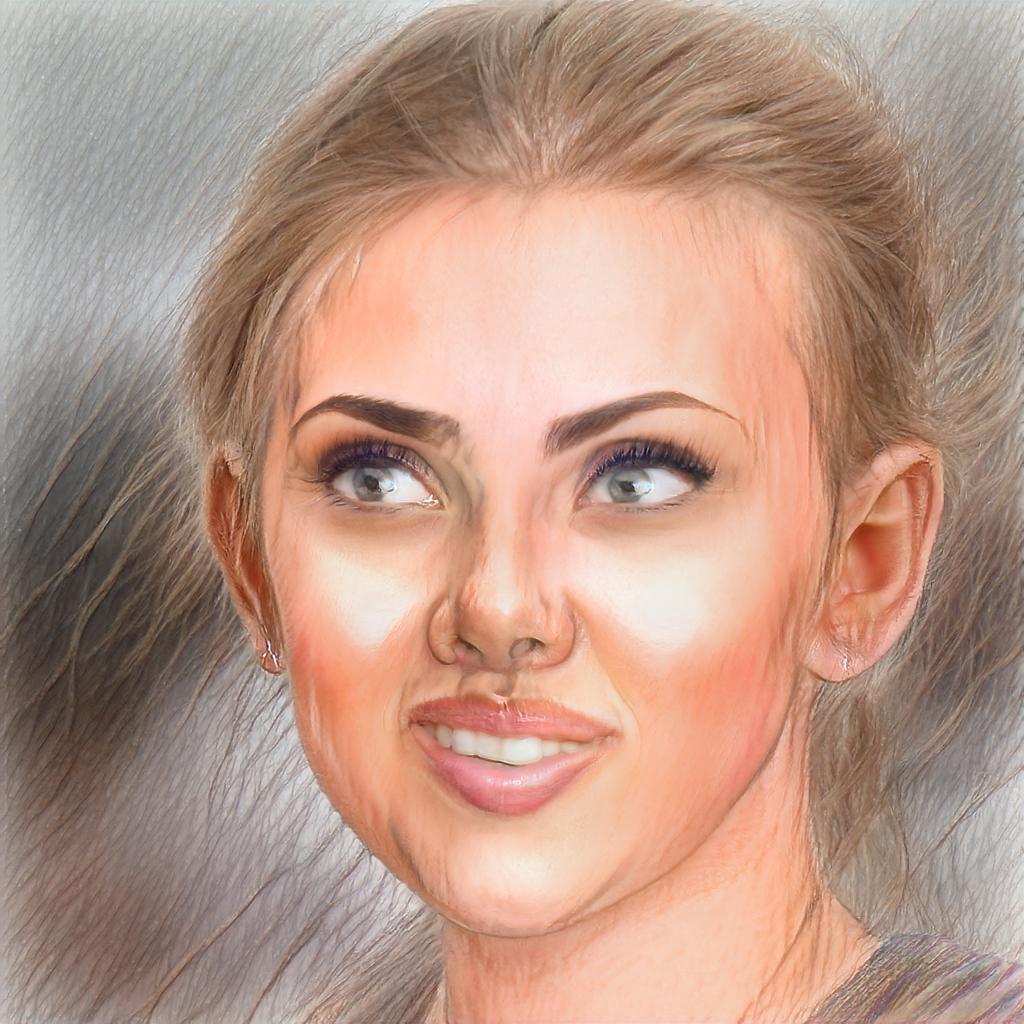} & 
  \includegraphics[width=0.32\columnwidth]{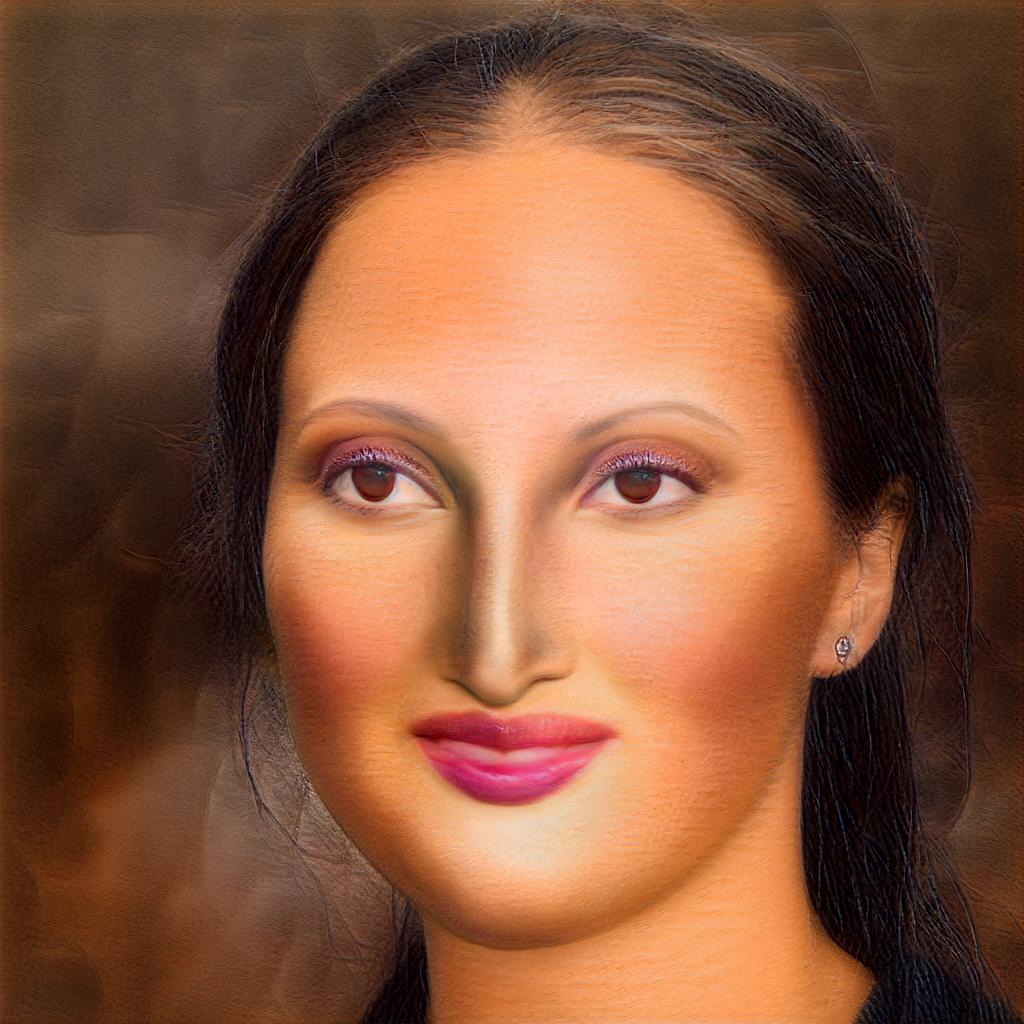} &
  \includegraphics[width=0.32\columnwidth]{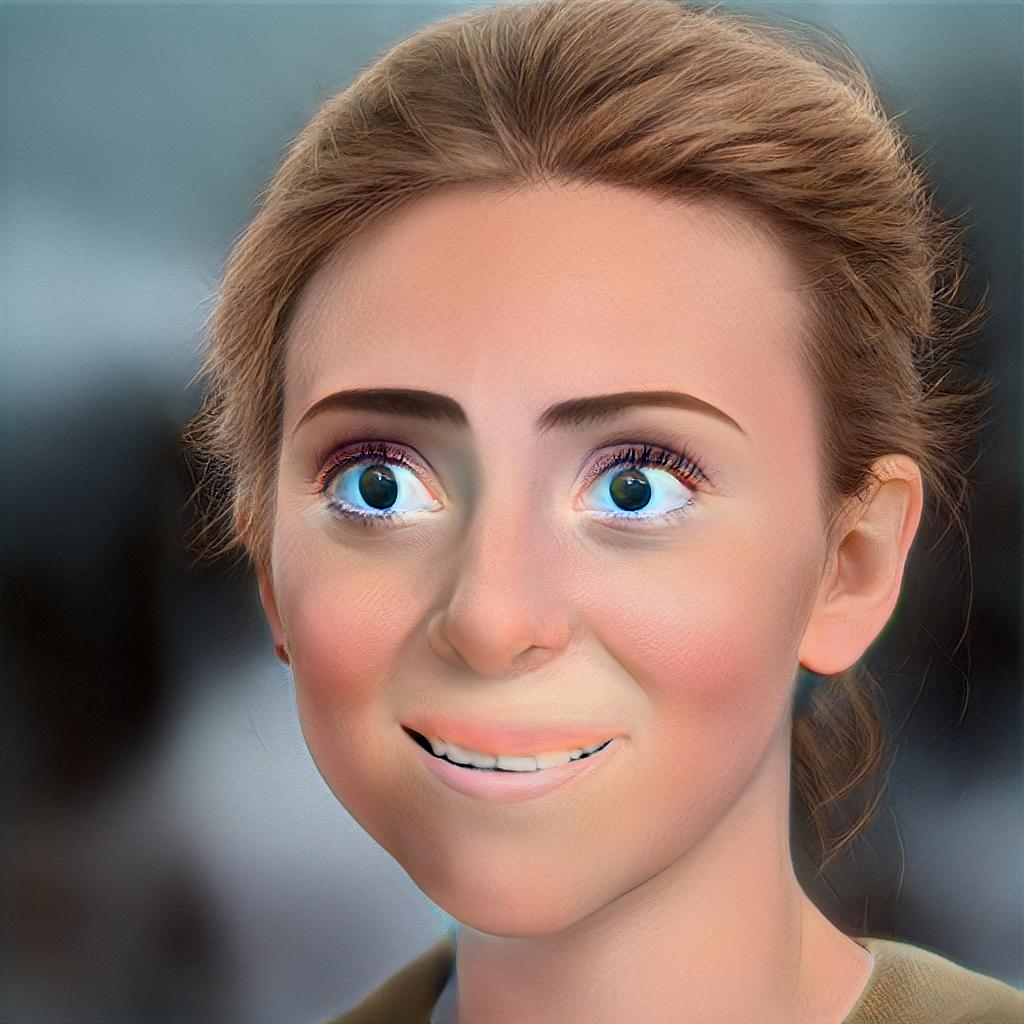}  
  \\
  Sketch & Mona Lisa & Pixar

\end{tabular}
\vspace{-0.25cm}
\caption{Editing a real image of Scarlett Johansson (on the top left) with StyleGAN. We show both in-domain and out-of-domain manipulations.}
\vspace{-0.7cm}
\label{fig:edit}
\end{figure}

StyleGAN presents a fascinating phenomenon. It is unsupervised, and yet its latent space is surprisingly well behaved. As it turns out, it is so well behaved that it even supports linear latent arithmetic. 
For example, it supports adding a vector representing age to a set of latent codes, resulting in images representing the original individuals, but older. Similarly, it has been demonstrated that StyleGAN arranges its latent space not only linearly, but also in a disentangled manner, where traversal directions exist that alter only specific image properties, while not affecting others. Such properties include global, domain-agnostic aspects (e.g., viewing angles or zoom), but also domain-specific properties such as expressions or gender for human faces, car colors, dog breeds, and more (see Figure~\ref{fig:teaser}, and Figure \ref{fig:edit}). Exploring what these qualities entail, recent StyleGAN-based work has presented astounding realism, impressive control, and inspiring insights into how neural networks operate.

As groundbreaking as it may be, these powerful editing capabilities only reside within the model's latent space, and hence only operate on images generated by StyleGAN itself. Seeking to bring real-world images to the power of StyleGAN's latent control, inversion into StyleGAN's latent space has received considerable attention. Further harnessing StyleGAN's powers, other applications have also arisen, bringing contributions to the worlds of segmentation, augmentation, explainability, and others. 

In this report, we map out StyleGAN's phenomenal success story, along with analyzing its severe drawbacks. We start by discussing the architecture itself and analyze the role it plays in creating the leading generative model since its conception in 2018. %
We then shift the discussion to the resources and characteristics StyleGAN's training requires, and lay out the work that reduces, re-uses, and recycles it.  

In Section  \ref{sec:editing}, we discuss StyleGAN's latent spaces. We show how linear editing directions can be found, encouraged, and leveraged into powerful semantic editing. We inquire into what properties StyleGAN can and cannot disentangle well and dive into a surprisingly wide variety of approaches to achieve meaningful semantic latent editing.

In Section \ref{sec:encoding}, the quest for applying StyleGAN's power in real-world scenarios turns to a discussion about StyleGAN inversion. To express a given real image in StyleGAN's domain, many different approaches have been suggested, all of which thoroughly analyze and exploit the generator's architecture. Some propose latent code optimization and others apply data-driven inference. Some works seek an appropriate input seed vector, while others interface with StyleGAN at other points along the inference path, greatly increasing its expressive power. Unsurprisingly though, it turns out that the well-behaved nature of StyleGAN's latent space diminishes in regions far from its well-sampled distribution. This in practice means that given a real-life image, its accurate reconstruction quality (or \textit{distortion}) comes at the cost of \textit{editability}. Finding different desired points on this reconstruction-editability trade-off is a main point of discussion in the works covered in this section.

Encoding an image into StyleGAN's latent space has more merit than for image inversion per se. There are many applications where the image being encoded is not the one the desired latent code should represent. Such encoding allows for various image-to-image translation methods \cite{Nitzan2020FaceID,richardson2020encoding,alaluf2021matter}. In Section \ref{sec:encoding}, we present and discuss such supervised and unsupervised methods. 

In Section \ref{sec:discriminative}, we show the competence of StyleGAN beyond its generative power and discuss the discriminative capabilities StyleGAN can be leveraged for. This includes applications in explainability, regression, segmentation, and more.

In most works and applications, the pre-trained StyleGAN generator is kept fixed. However, in Section \ref{sec:fine-tuning}, we present recent works that fine-tune the StyleGAN generator and modify its weights to bridge the gap between the training domain (in-domain) and the target domain, which could possibly be out-of-domain. 

Each section addresses both the newcomer, with basic concepts and conceptual intuition, and the experienced, with a summary of the most established and promising approaches, along with some pointers regarding when to use them.

\section{StyleGAN Architectures}
\label{sec:architecture}

This report addresses the benefits hidden in Generative Adversarial Networks (GANs). First introduced by Goodfellow~\etal~\shortcite{Goodfellow2014GenerativeAN}, GANs pose an interesting and unique approach. Two networks are interlocked in a perpetual game during training. One network, the Generator, seeks to generate images that are from the target distribution, while the other network, the Discriminator, seeks to distinguish between actual images from the training set and those created by the generator. The two networks start the training without any knowledge of the domain and spend the entire training process learning from each other. Conceptually, this could be thought of as a repetitive process where the generator finds a way to fool the discriminator, and the discriminator, in turn, finds a way to detect this "attack". This approach allows self-supervision, and hence these networks can be trained without explicit labeling.

StyleGAN, however, seems to do much more than reproduce samples from the target distribution. While following the adversarial learning process, it turns out that StyleGAN, more than other GANs, constructs a remarkably well-behaved latent space. Without any supervision, StyleGAN arranges examples it sees in a smooth, highly disentangled order, driven by powerful semantic understanding. 

In this Section, we portray how StyleGAN's architecture is built, try to understand why this architecture induces such cutting-edge emerging disentanglement, and how the architecture can be improved to match specific needs, according to relevant literature.

\paragraph*{StyleGAN1} The style-based generator architecture for generative adversarial networks, or StyleGAN for short, was first proposed by Karras~\etal~\shortcite{karras2019style}. At the core of StyleGAN's architecture lie the style modulation layers, from which StyleGAN draws its name. Borrowing from style-transfer literature, these layers are designed to enable control over the ``style" of generated images by adjusting the statistics of the feature maps along the generative path. The generative path starts from a learned constant $C$, representing the epicenter of the distribution, and all the information and generative power of the network is injected through the style and an additional random noise vector $n$. In the first version of the architecture, \cite{karras2019style}, the style injection layers utilized the Adaptive Instance Normalization (AdaIN) mechanism~\cite{huang2017arbitrary}; each channel of the feature maps is first normalized to zero mean and unit variance, followed by re-scaling using new means and variances predicted from a given latent code. 

However, the use of AdaIN layers was not the only major change proposed. Rather than injecting the network with a latent code $z$ sampled directly from some Gaussian prior $\mathcal{Z}$, StyleGAN introduces a novel mapping network which converts these
\begin{wrapfigure}{r}{0.25\linewidth}
\hspace*{-0.8cm}  
\centering
\includegraphics[width=0.95\linewidth]{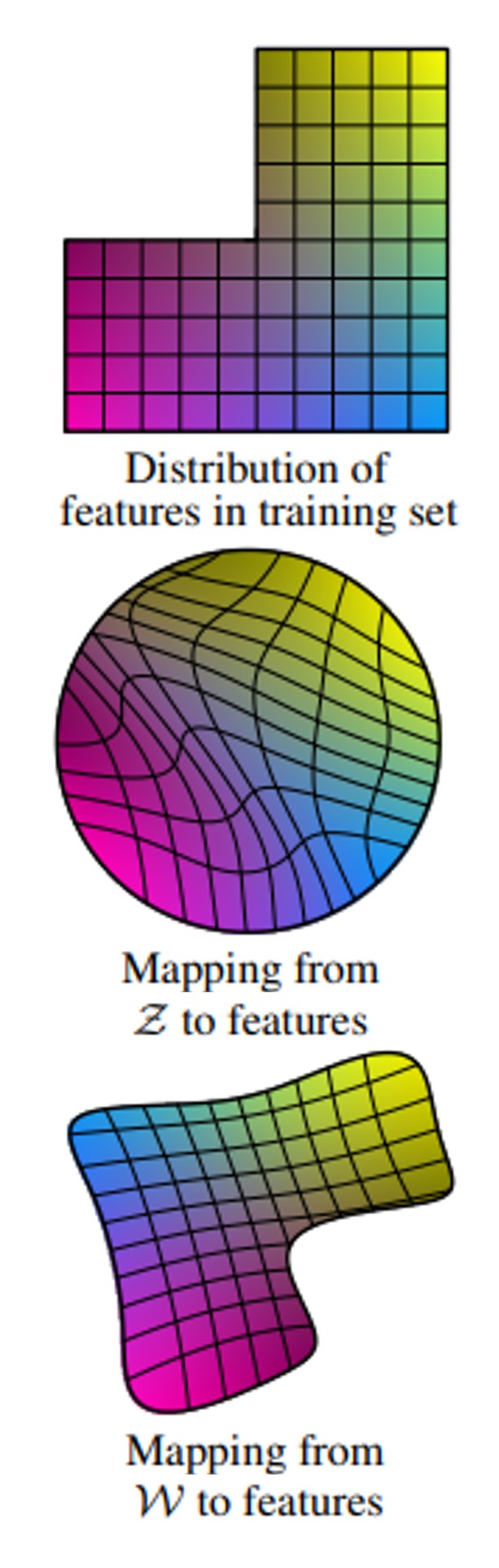}
\end{wrapfigure}
normal-distributed codes into vectors in an intermediate latent space $\mathcal{W}$. The authors propose an intuitive argument for adding such a network: the probability for sampling a particular combination of image attributes in the latent space should eventually match the probability for that combination to appear in the real dataset. For those cases where the data is not uniform with respect to these attributes, it follows that the mapping from $\mathcal{Z}$ to the image features must become curved in order to diminish the incidence rate of rare attribute combinations.
A learned mapping network, however, could learn to ``unwrap" the latent space back to a flat form, and simply account for probability densities by mapping fewer codes to regions that would otherwise portray a rare combination of attributes (see inset figure, from Karras~\etal~\shortcite{karras2019style}). Karras~\etal postulate that this linearly-disentangled space is a more natural representation for the network, allowing it to more easily recreate a wide range of variations. As Karras~\etal and follow-up works demonstrate, the learned latent spaces of StyleGAN offer considerable disentanglement. These innovations give rise to a network that, at the time, was unrivaled in quality, invertibility, and support for a wide range of generative and discriminative tasks. 

\begin{figure}
\includegraphics[width=0.99\linewidth]{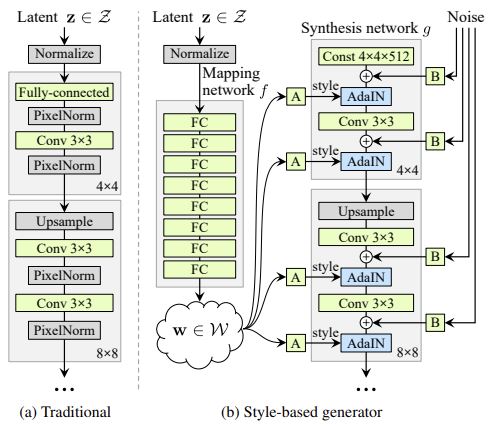} 
\caption{The StyleGAN1~\cite{karras2019style} architecture. The novel architecture is based on the progressive growing approach (b, right), combined with a Style injection mechanism (b, middle). In addition, a mapping network (b, left) deforms the Gaussian $Z$ space to better match the distribution of the training data.}
\vspace{-2pt}
\label{fig:sg1_arch}
\end{figure}

\paragraph*{StyleGAN2} With StyleGAN being quickly adopted into widespread use, it was inevitable that artifacts inherent to the model would come to light. These included characteristic water-droplet-shaped blobs which consistently appeared in all images. Additionally, a ``texture-sticking" effect was observed, where certain attributes of the generated image, such as the teeth or eyes, would display strong spatial bias, remaining fixed to specific image coordinates even through latent space interpolations. In a follow-up work, Karras \etal \shortcite{karras2019analyzing} identify the source of these artifacts, and re-design key aspects of the network to correct them.
They first identify that the water-droplets are a manifestation of a flaw in StyleGAN's normalization scheme - by normalizing each feature map separately, any information found in their relative magnitudes is destroyed. The generator would then hide information about the signal strength through localized spikes that dominate the statistics. They overcome this hurdle by shifting normalization from the feature maps, where the adaptive layer normalization forces new statistics, and onto a modulation of the convolutional kernels themselves. By doing so, they apply weaker normalization, based on the expected feature statistics rather than exact signal strength, and the network no longer needs to hide signal strength information - which in turn makes the blob-shaped artifacts disappear. This technique has also been shown to promote disentanglement between geometry and appearance in other scenarios~\cite{yang2020iorthopredictor}.

The ``texture sticking" effect, meanwhile, was hypothesized to be an artifact of the progressive growth scheme. Karras \etal suggest that in such a setup, every resolution block serves as an output block for some stage of the training process. In such a scenario, the network attempts to create excessive high-frequency detail in these intermediate resolutions, which leads to aliasing along the generative path and in turn breaks shift-invariance \cite{zhang2019making}. They address this issue by revisiting progressive growing and replacing it with a skip-connection-based architecture, where each resolution block outputs a residual, which is summed up and up-scaled. These modifications, coupled with a novel path-length regularization loss and in-depth analysis of network capacity, lead to improvements both in standard quality metrics such as FID, but also in the ability to invert images into the latent space of the GAN.

\begin{figure}
\includegraphics[width=0.995\linewidth]{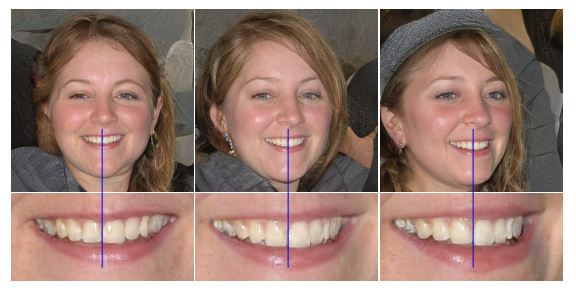} 
\vspace{-0.3cm}
\caption{An example for the ``texture sticking" effect~\cite{karras2019analyzing}. As can be seen, the teeth do not follow the head when rotated, but rather remain attached to their absolute position in the image. }
\vspace{-2pt}
\label{fig:texture_sticking}
\end{figure}

\paragraph*{StyleGAN3} At first, StyleGAN2 appeared to address the ``texture-sticking" problem. However, more careful analysis revealed that, while the issue was resolved for large-scale objects such as the mouth or the eyes, it remained present when examining finer details such as hair or beards. To resolve this issue, Karras \etal sought out the various sources through which spatial information could leak into the convolutional operations, with the aim of fully restoring translational invariance to the network. These sources include the image borders, per-pixel noise inputs, positional encoding, and the aliasing caused by careless treatment of upsampling filters and non-linearities such as ReLUs. Through a series of small architectural changes coupled with a rigorous signal processing approach, these sources of unwanted information were removed, and translation and rotational equivariance was restored. The novel architecture of StyleGAN3~\cite{karras2021alias} brought with it remarkable improvements, leading to considerably smoother interpolations. However, the new approach brought with it new challenges. Karras \etal observe that when conducting layer mixing experiments, some properties were not cleanly inherited from just one of the codes. Preliminary investigations of the network also revealed newly introduced artifacts, from the tendency of generated faces to have a single frontal tooth, to the appearance of a faint ``grid" to which background features and fine details such as hair would often get stuck. These phenomena suggest degraded disentangled properties, however, as of writing these words, the novel alias-free architecture is still in its infancy, and it remains to be seen what unique uses, improvements or challenges arise from it.

Parallel to these improvements, various works sought to identify areas in which StyleGAN could be improved. Lin \etal~\shortcite{lin2021anycost} note that the high computational cost of full-resolution image generation makes it impractical to utilize the network for interactive editing on edge devices. They proposed an elastic generator architecture that could produce previews at lower resolutions while retaining the same latent semantics. A user could then edit these previews with a fraction of the computational budget, restoring the output to its full resolution only as a final step.
\cite{swagan2021gal} followed prior observations which revealed a flaw in the generator's ability to produce high-frequency details. They demonstrated that some patterns are beyond the network's ability to recreate and linked the flaw to the inherent spectral bias of neural networks. They proposed to tackle this by shifting the generation to the frequency domain, realized by a first-level wavelet decomposition. By doing so, they reduced the network's need to learn high-frequency functions and achieved a more faithful generation of high-frequency patterns. 

\begin{figure}
\includegraphics[width=0.99\linewidth]{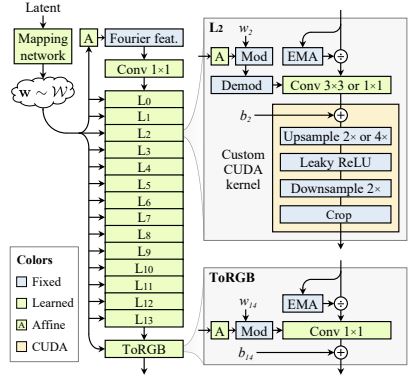} 
\vspace{-0.15cm}
\caption{StyleGAN3~\cite{karras2021alias} architecture. The main components of the architecture remain similar to previous versions. A series of small architectural changes, derived from rigorous signal processing analysis, renders the new version of StyleGAN equivariant to rotation and translation. }
\vspace{-0.35cm}
\label{fig:sg3_arch}
\end{figure}

In an alternative approach to synthesis, Anokhin \etal~\shortcite{anokhin2020image} forgo convolutions and instead design a style-based network which, given the coordinates of a pixel and a style code, predicts the color of that pixel. This conditionally-independent pixel synthesis approach (CIPS) was able to rival the quality of images produced by traditional convolutional methods while enabling novel synthesis applications such as the creation of cylindrical panoramas.
Sendik \etal~\shortcite{sendik2020unsupervised} hypothesize that the single learned constant at the root of the generative path is a limiting factor when training on sets that contain multiple modalities. They hence develop a multi-constant model, where the generator could better represent the dataset modalities by assigning them different mixtures of constants.
Kwon \etal~\shortcite{kwon2021diagonal} propose augmenting the network with Diagonal Spatial Attention (DAT) layers, which modulate the network's feature maps along the spatial directions. These modulations are in turn controlled through an additional latent code. Through this addition and an appropriate loss term, the authors disentangle ``content" from ``style", allowing a user to control spatial features such as pose or expression, without affecting style traits such as color or makeup.
Casanova \etal~\shortcite{casanova2021instanceconditioned} suggest training a GAN conditioned on a single input. Their intuition is that unconditional GANs face difficulties in reproducing complex distributions \cite{luvcic2019high,liu2020diverse} such as ImageNet \cite{russakovsky2015imagenet}. Typical conditional models seek to resolve this challenge \cite{brock2018large} by conditioning the synthesis process on class labels, thereby partitioning the data into multiple clusters which are more easily modeled. However, acquiring such labels is labor intensive. Instead, they suggest partitioning the data into overlapping neighborhoods by clustering the data in some pre-trained feature space. The ``label" associated with an image is then the feature vector in this space, and real images observed by the discriminator when conditioned on such a vector are sampled from the group of images with representations most similar to the given vector. In this way, the network learns to generate images sharing visual and semantic traits with a given sample.

While not strictly extensions of StyleGAN itself, a large body of work nevertheless draws inspiration from its novel architecture. These works typically repurpose the style-based modulation layers or mapping network and incorporate them into new generative frameworks. One line of work aims to merge the growing Transformer~\cite{vaswani2017attention} literature with image synthesis. Hudson \etal~\shortcite{hudson2021generative} proposed the Generative Adversarial Transformer, which utilizes a bipartite mechanism through which the latent codes and image features attend and influence each other. 

Others have proposed to entirely replace the convolutional blocks with transformer-based modules such as ViT~\cite{lee2021vitgan,dosovitskiy2020image}, Linformer~\cite{park2021styleformer,wang2020linformer}, or the Swin Transformer~\cite{liu2021swin,zhang2021styleswin}. While these have yet to achieve the same fidelity or widespread use as their progenitor, they have already shown considerable progress in layout control and convergence times.
Moving towards 3D representations, a set of recent works propose to marry the style-based architecture with implicit models, such as Signed Distance Functions~\cite{park2019deepsdf} or Neural Radiance Fields~\cite{mildenhall2020nerf}. These models leverage weight and feature modulations~\cite{gu2021stylenerf,orel2021stylesdf,zhou2021cips,xu2021volumegan} or directly employ a StyleGAN network to predict a set of feature planes that serve as inputs to a small implicit network~\cite{Chan2021eg3d}. These works achieve impressive visual quality, enable explicit control over pose, and can be used to predict detailed surface representations. However, their increased memory requirements have so far prevented them from reaching the resolution and quality of StyleGAN itself.

\paragraph*{Training Data} "An open secret in contemporary machine learning is that many models work beautifully on standard benchmarks but fail to generalize outside the lab"~\cite{jahanian2019steerability}. Indeed, StyleGAN is no different. It is recognized in the literature that unsupervised training is more difficult when learning a complex domain~\cite{casanova2021instanceconditioned}. In the case of StyleGAN, the learned domain seems to require strict structure. The data domain should be almost convex, i.e., between every two points there should be valid samples that interpolate them on the data manifold. For this reason, for example, it is difficult to construct a full human body model. For the same reasons, StyleGAN does not handle multi-modal distributions well and behaves poorly for scenes where objects do not have specific potential locations. In recent work, Sauer~\etal~\shortcite{sauer2022styleganxl} demonstrate that some of these challenges may be overcome through careful model scaling, though whether or not StyleGAN's unique latent-space properties persist through this modification remains an open question. In the future, we will likely witness additional works that address explicit data issues, i.e., works that try to apply StyleGAN to other types of data, perhaps by dropping or adding examples during training to make the data's landscape more smooth, by transfer learning between datasets (see Section~\ref{sec:fine-tuning}), by more directly addressing multi-modalities in the data, or by incorporating more elaborate attention mechanisms into the architecture. 
\begin{figure*}[!ht]
    \centering
    \includegraphics[width=1.01\textwidth]{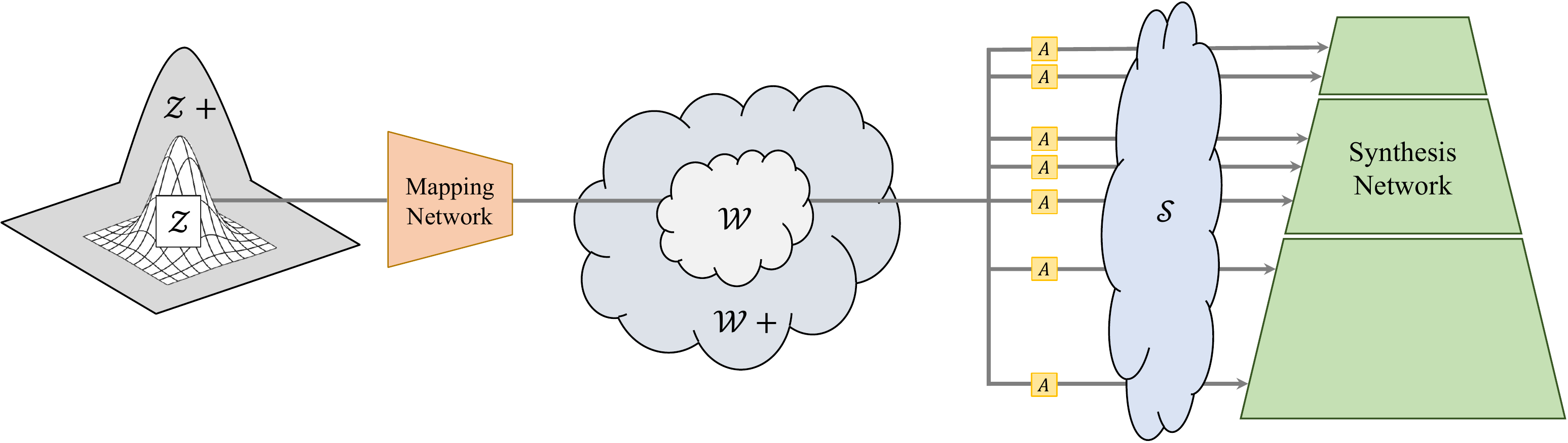}\vspace{11pt}
    \caption{The StyleGAN architecture and its latent spaces. A random latent code $z$ is sampled from the normally distributed latent space $\mathcal{Z}$ (on the left), then transformed to the learned latent space $\mathcal{W}$ through an MLP mapping, passed through a set of different learned affine transformations (denoted by $A$) to reach the $\mathcal{S}$ space, and finally inserted into the synthesis network. It is common to work in the extended spaces of $\mathcal{Z}$ and $\mathcal{W}$, referred to as $\mathcal{Z+}$ and $\mathcal{W+}$, respectively. } \vspace{5pt}
    \label{fig:space}
\end{figure*}

\vspace{-0.2cm}
\subsection{Latent Spaces}
\label{sec:spaces}

Unlike common GANs, StyleGAN has more than one innate latent space. Moreover, to increase the expressive power of StyleGAN, it is common to work with extensions of these spaces, illustrated in Figure~\ref{fig:space}. Here, we review the commonly used spaces and describe the differences between them. 

\begin{itemize}
    \item
    The first latent space is $\mathcal{Z}$ in the sense that random latent codes can be sampled from it to be inserted into the generator itself. $\mathcal{Z}$ is defined to be a normally distributed space, and it is the only space that has a closed-form definition. Therefore, images that belong to the GAN's manifold can be easily sampled from $\mathcal{Z}$.

    \item
    Latent codes from $\mathcal{Z}$ are transformed to latent codes in $\mathcal{W}$ through an MLP, commonly referred to as the \textit{mapping network}. In a sense, the distribution of $\mathcal{W}$ is learned, and therefore better matches the distribution of the real data compared to the original $\mathcal{Z}$ space. This learned distribution provides the virtue of disentanglement. Many works employ this disentanglement property to achieve semantic image editing by traversing the latent space. 

    \item
    Latent codes in $\mathcal{W}$ are not directly inserted into the synthesis network. Instead, each latent code in $\mathcal{W}$ is first transformed through a learned affine transformation. Such an affine transformation is learned during the training of each layer of the synthesis network. 
The space spanned by the outputs of these transformations is commonly referred to as the \textit{StyleSpace}, or $\mathcal{S}$. Unlike $\mathcal{W}$ in which a single latent code is used for generating an image, in $\mathcal{S}$ there are several latent codes for a single image, one for each affine transformation block (e.g., 26 for a generator with a $1024 \times 1024$ output resolution). It has been shown \cite{wu2020stylespace} that $\mathcal{S}$ is even more disentangled than $\mathcal{W}$. More specifically, each dimension, or channel, of $\mathcal{S}$ tends to control a single semantic attribute of the generated image. Therefore, by carefully manipulating the dimensions of $\mathcal{S}$ it is possible to obtain highly disentangled edits.

    \item
    Representing real images with StyleGAN remains a challenge. The good properties of $\mathcal{W}$ have attracted most works aiming at representing real images to focus on it. Abdal~\etal~\shortcite{abdal2019image2stylegan} propose working in an extended latent space, denoted by $\mathcal{W+}$
    In $\mathcal{W+}$, one inserts a different latent code for each layer of the synthesis network (e.g., $18$ for a generator with a $1024 \times 1024$ resolution). StyleGAN was not trained on $\mathcal{W+}$ and thus images sampled from it do not necessarily have high quality. Moreover, it should be noted that oftentimes, when operating in $\mathcal{W+}$, it is possible to reach areas that are outside the learned distribution of $\mathcal{W}$. Such areas further push the latent code outside the distribution over which the generator was trained on. As the distribution of $\mathcal{W}$ cannot be explicitly modeled, keeping the latent code in the trained distribution is a challenging task.
    \item
    To alleviate the need of preserving the latent code inside the distribution of $\mathcal{W}$, it is possible to work with an extension of $\mathcal{Z}$ instead of $\mathcal{W}$. Similarly to the definition of $\mathcal{W+}$, in $\mathcal{Z+}$ \cite{song2021agilegan} a different latent code is sampled for each layer of the synthesis network (e.g., $18$ for a $1024 \times 1024$-resolution generator).
    Note, that in $\mathcal{S}$ there is no notion of $\mathcal{S+}$ as the latent codes for each layer are different by design.
\end{itemize}
\section{Latent Space Editing}
\label{sec:editing}

Perhaps the most exciting aspect of GAN learning is the way the latent space is arranged in a well-trained GAN. Traditionally, GANs in general, and StyleGAN specifically, can be used to simply generate a wide variety of images of the same kind. These can serve as a form of data augmentation for downstream training (see Section~\ref{sec:discriminative}). However, it has been shown that GANs tend to arrange their latent space smoothly, \ie{} such that close regions in the latent space depict similar images. 

This, combined with the notion that GANs produce images that are within the distribution of the target domain gives rise to \textit{latent-based editing}. In other words, the two concepts suggest that traversing the latent space yields a path of smoothly changing images, each of them on their own belonging to the target domain (\eg{} realistic human faces). This could be thought of as geodesic traversal on the manifold of all valid images. Even the first works in generative modeling already demonstrated how latent code interpolation between two examples yields a natural morphing between them \cite{Goodfellow2014GenerativeAN}. As it turns out, careful traversal in the latent space can also produce desirable semantic changes in the resulting image that would otherwise be very difficult to perform. These include changes in viewpoint, lighting conditions, and domain-specific attributes such as expressions for faces, colors for cars, or widths of buildings. Of course, the most desirable edits are the disentangled ones --- those that change one attribute without affecting any other. Applications of such powerful editing tools are endless, from automatically adding smiles to facial images, through interior design explorations, to rapid car design.

In this aspect, StyleGAN shines. As previously discussed (Section~\ref{sec:architecture}), StyleGAN operates best on well-structured data. When trained on such data, StyleGAN constructs a highly disentangled latent space in an unsupervised manner, simply by virtue of inductive bias. Many techniques have been proposed to traverse this latent space and facilitate semantically disentangled latent-based editing. Of all sections in this report, the editing art is the most diverse, presenting creative approaches borrowed from different fields. 

Early approaches to this task pointed out that StyleGAN's latent space is so well behaved and disentangled, that it even supports linear latent space arithmetics. These linear editing works demonstrate, for example, that to make a face older, one can traverse in a specific, pre-computed, direction. 
These works come in two main flavors --- supervised and unsupervised. The first works in the field have presented a thorough analysis of GAN behavior \cite{jahanian2019steerability,liu2020style} (including StyleGAN), and showed how one can identify linear traversal directions that present high disentangled qualities. Using edits that are easily attainable in image space (e.g. 2D rotation or zoom, and pan), they look for directions in the $\mathcal{W}$ space (Section~\ref{sec:spaces}) that produce the same effect. Changing the magnitude of traversal along these directions induces a disentangled edit that is weaker or stronger according to the step size. This early work also drew conclusions regarding the extent of the space's linearization. That is, they show that going too far along a direction will eventually break the disentanglement, and affect other crucial factors of the image. They also offer an analysis on, and a way to improve, the extent of the linearization. 

\begin{figure}[!t]
  \centering
    \includegraphics[width=\columnwidth,keepaspectratio]{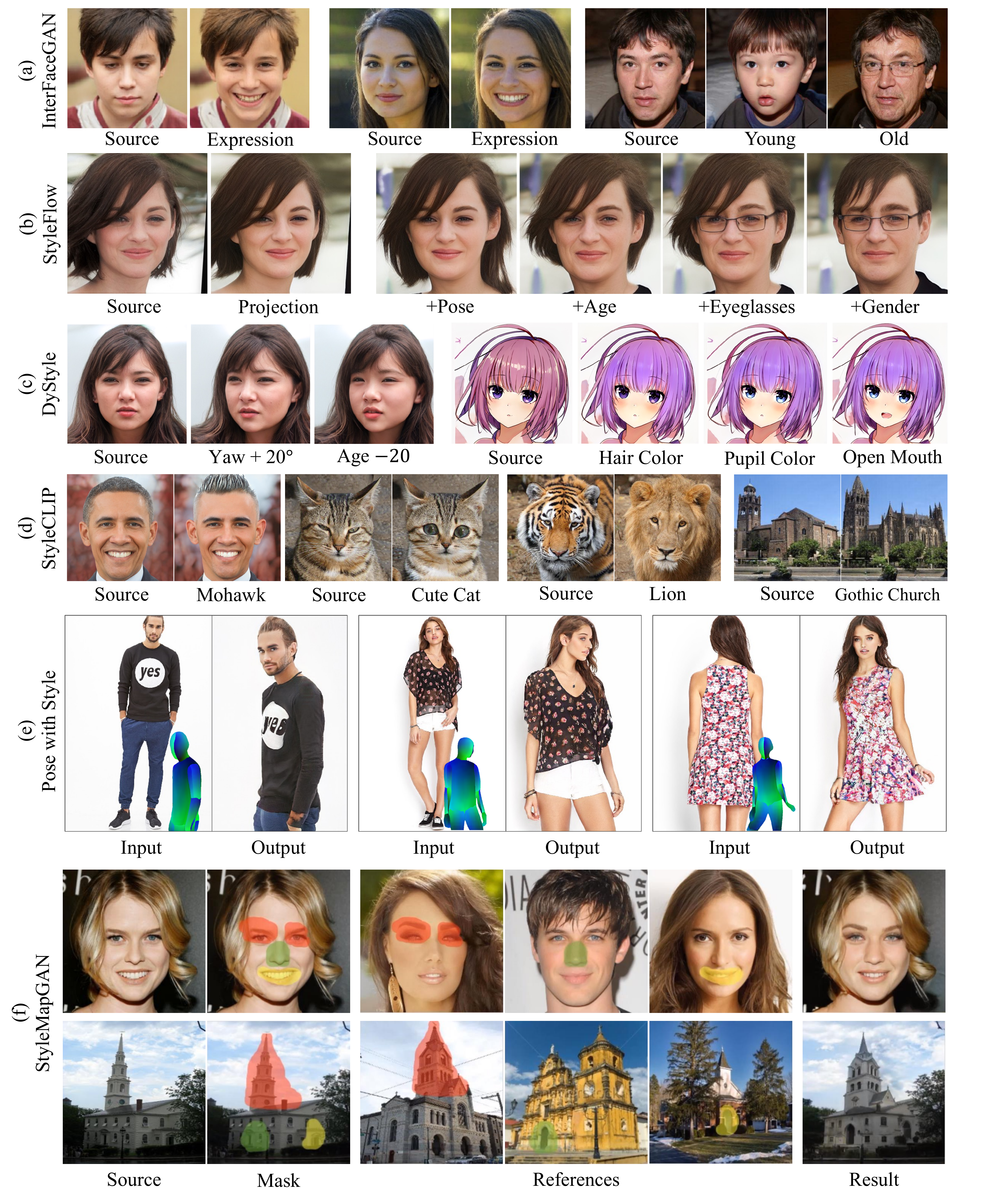}
  \caption{Examples of prominent editing works. (a) InterfaceGAN~\cite{shen2020interfacegan} extracts linear editing directions through attribute level supervision. (b) StyleFlow~\cite{abdal2020styleflow} is the first to present editing that is stable enough to be composed, through employing normalizing flows and attribute-level supervision. (c) DyStyle~\cite{li2021dystyle} addresses compositional editing directly, producing more accurate, elborate, and diverse editing. (d) StyleCLIP~\cite{patashnik2021styleclip} employs free textual editing, through a visual-linguistic pretrained model \cite{radford2021learning}. (e) Pose with Style~\cite{albahar2021pose} employs human pose supervision to edit body poses and clothing. (f) StyleMapGAN~\cite{kim2021stylemapgan} provides localized editing by augmenting StyleGAN's architecture with spatially adaptive modulation. Zoomed-in viewing recommended. }
  \label{fig:edits}
  \vspace{-2pt}
\end{figure}

\vspace{-0.2cm}
\paragraph*{Supervised Linear Approaches} The most natural approach to finding editing directions is to do so explicitly, through full supervision. Perhaps one of the most noteworthy works in linear editing is InterFaceGAN~\cite{shen2019interpreting,shen2020interfacegan}. %
This work leverages per-image binary annotations to identify hyper-planes in the latent space that separate the two binary attribute values. These planes can be found using Support Vector Machines (SVMs). Then, to edit one attribute without affecting others, one finds a direction that is orthogonal to one plane and parallel to the others. Figure~\ref{fig:edits}a depicts some of the typical editing directions extracted by this method. Yang~\etal~\shortcite{yang2020semantic} further propose a way to evaluate how well the activations of specific layers are correlated with semantic attributes, based on 105 pretrained attribute classifiers. Recently, Wu~\etal~\shortcite{wu2020stylespace} employ a pretrained classifier, or use a few images for direction identification. Their key idea is to identify correspondence between the most active channels and the semantics corresponding generated images depict (termed \textit{semantic consistency}). The authors show that this correspondence indicates specific channels in the generator's activations that control very disentangled image characteristics. This offers a fine-grained approach to latent editing that is different from the popular latent-editing approaches which modify all activations of a layer (or more). Editing in $\mathcal{S}$ space (see Section~\ref{sec:spaces}) is shown to provide highly disentangled, spatially adaptive directions for editing.

\paragraph*{Unsupervised Linear Approaches} 
In many cases, collecting the data required for supervised editing can be difficult or prohibitively expensive. To expand the range of available editing directions, despite these limitations, unsupervised editing methods have been proposed. Perhaps the first was proposed by Voynov \etal~\shortcite{voynov2020unsupervised}. The core idea is to predict a set of traversal directions and concurrently try to infer their meaning from the images corresponding to the code before and after the edit. They propose to jointly learn a set of directions and a model to identify the corresponding image transformations. Under this paradigm, the assumption is that directions that are easy to identify with high accuracy are likely candidates for disentangled editing directions. GANSpace \cite{harkonen2020ganspace}, take a more natural approach, and simply search for the dominant directions in the latent codes of a dataset, using Principal Component Analysis (PCA). Alharbi~\etal~\shortcite{Alharbi_2020_CVPR} propose editing through adding random noise to the input learned constant, rather than augmenting the style input. They show that by enforcing a spatial structure to the noise, spatial disentanglement can be encouraged, and can be paired with the semantic disentanglement StyleGAN already offers. In all three cases, manual inspection is used to identify whether these directions indeed produce valuable edits, and infer their semantic meaning. SeFa~\cite{shen2020closedform} takes a different approach to the unsupervised editing problem. They propose analyzing the weights of the pretrained generator and identifying principle directions that are most affected by these weights. To do so, they perform an eigenanalysis of the matrix representing the latent-to-image space projection. This analysis is closed-form, meaning it is fast and does not require even sampling the network. This approach is still valuable and has been used for other domains and GAN architectures as well ~\cite{spingarn2020gan}. 

\paragraph*{Non-linear Approaches}
As may be expected, non-linear approaches can present higher quality editing at the cost of simplicity.
Hou~\etal~\shortcite{hou2020guidedstyle}, operate similarly to Yang \etal~\shortcite{yang2020semantic} by using classifiers. However, they propose to move beyond global, linear directions and towards a non-linear traversal paradigm. In their case, a different direction is generated per example for the same editing operation. The editing is then performed by changing the latent code of only one layer at a time, in a style-mixing manner (see Section~\ref{sec:architecture}), thereby improving disentanglement. StyleFlow~\cite{abdal2020styleflow} is a seminal work in the realm of facial editing, presenting one of the most versatile and stable editing approaches, disentangled enough to produce a realistic result even when performing several editing operations serially, as can be seen in Figure~\ref{fig:edits}b. The core idea for this work is the clever employment of normalizing flows --- a method through which a bi-directional mapping can be obtained between the latent space and an input code, conditioned on specific attributes. This mapping is trained in a supervised manner through an elaborate multi-attribute classifier. This promising normalizing flow-based approach has also seen follow-up work in an unsupervised setting~\cite{liang2021ssflow}. Alaluf \etal~\shortcite{alaluf2021matter} use an age regression network to provide control over age in human faces. Looking towards more recent works along this line~\cite{wang2021hijack}, perhaps the state-of-the-art lies with DyStyle~\cite{li2021dystyle}. The main contribution of this supervised approach is a dynamic network, trained to handle multiple edits at the same time. Here, a different network is trained for each attribute, producing its own latent editing direction. For every training example, consisting of a different composition of desired edits, only the relevant networks are applied, with their outputted codes fused into one using a self-attention mechanism. This approach enables high-quality editing in flexible domains, especially when composing several edits together. The combined dynamic approach seems not only to improve sequential editing, but also provide enough regularization to improve the state-of-the-art for a single edit as well (see Figure~\ref{fig:edits}c). 
Aiming for video editing, Yao \etal~\shortcite{yao2021latent} train a dedicated latent-code transformer to achieve more disentangled edits.

\paragraph*{Different Supervision Modalities} 
Other approaches have been proposed that leverage supervision, but differ in nature from explicit attribute supervision or classification-based techniques. StyleRig~\cite{tewari2020stylerig} suggests employing synthetic data to guide the editing process. They acquire a roughly 200-parameter 3D Morphable Face Model (3DMM) using traditional PCA over 200 input faces. This model can be used in a self-supervised manner to train a network to perform the editing over \w{}. Through a plethora of synthetically generated paired examples, the method finds high-quality edits.  This is because perfect labeling can be assigned to images that are rendered by specific parameter changes in the 3DMM model. This approach, however, was only able to find high-quality editing directions for a subset of the face model parameters. Perhaps unsurprisingly, the successfully found directions do not enable more diverse edits compared to less supervised methods. A similar approach has also been proposed~\cite{zhang2020image}, using general meshes instead of 3DMM, for more diverse objects. Through differentiable rendering, parameters like camera position and object shape can be self-supervised easily. Taking this line of work a step further, Ghosh~\etal~\shortcite{ghosh2020gif} propose generating the parameters of a 3D facial model learned from 4D scans. In this paradigm, the geometry is constructed through a learned 3D model (FLAME~\cite{li2017learning}), and StyleGAN generates appearance and texture. Combining the two models offers more expressive facial variations in shape and expression, and an inherent disentanglement between geometry and appearance.  FreeStyleGAN~\cite{leimkuhler2021freestylegan} use standard calibration tools to construct pairs of facial images and associated camera parameters. These pairs are used to learn explicit control over image views within the GAN's aligned image manifold. Taking this approach a step further, the authors use a flow-based model to learn an image mapping module that can transform the generated images beyond StyleGAN's aligned domain. HistoGAN~\cite{afifi2021histogan} employs color histograms to recolor images and paintings. 

Several works employ the power of language. They guide edits by using textual descriptions, which are more global and abstract in nature. Patashnik~\etal~\shortcite{patashnik2021styleclip}, one of the first works to propose this approach, employs CLIP~\cite{radford2021learning}, a powerful pre-trained model that embeds text and imagery to a joint latent space. By finding traversal directions that bring the produced image and the desired text description closer together, this method demonstrated new and exciting semantic editing operations, such as makeup removal and specific hairstyles for human faces (see Figure~\ref{fig:edits}e). TediGAN~\cite{xia2021tedigan} employ a novel architecture and training process for the language model to be trained along with the generator. While potentially powerful, the resulting networks are not as expressive as language models pre-trained on web-scale data. Hence, they fail to achieve the same quality. Chefer \etal~\shortcite{chefer2021image} utilize CLIP to blend two facial images, demonstrating better preservation of the original identity while successfully transferring meaningful semantic features from the desired target images. Abdal~\etal~\shortcite{abdal2021clip2stylegan} find meaningful directions in CLIP-space in an unsupervised manner, map them to latent-space directions, and use CLIP to automatically generate natural language descriptions for these directions.

\paragraph*{Finer Control} Several of the latest works propose operating in a more disentangled latent space --- the $S$ space~
\cite{wu2020stylespace,liu2020style,xu2021generative} (see Section~\ref{sec:spaces}). However, it is significantly larger, posing a computational challenge. Furthermore, augmenting the generator activations themselves after the AdaIN (StyleGAN~\cite{karras2019style}) or Modulation (StyleGAN2~\cite{karras2019analyzing}) layers, provides even finer control. This allows applying local changes in the image maps, rather than a global change. For example, Bau~\etal~\shortcite{bau2021paint} offers users the ability to paint a mask in a given image and to describe in free text what this region of the image should depict. They do this by feeding the same modulation layer different style codes, according to the spatial location in the resulting map: one code for regions inside the painted mask, and one for the rest.

Albahar \etal~\shortcite{albahar2021pose} suggest spatial control through the initial input constant, while leveraging the inherent semantic understanding StyleGAN naturally develops, reinforced by human pose labeling. Unlike most editing works, which manipulate the behavior of a pretrained StyleGAN, this work proposes architectural changes to the generator, to adapt it to human pose inputs. Through full supervision~\cite{cao2019openpose}, they train StyleGAN to change the pose of human clothing models. Through pose labeling and paired UV coordinates, the clothes are warped in UV space to better match the new pose (see Figure~\ref{fig:edits}c). 
Similarly, Abdal \etal~\shortcite{abdal2020image2styleganpp} change the spatial activations to allow scribble-level control for the user (see Section~\ref{sec:encoding} for more details). StyleFusion~\cite{kafri2021stylefusion} propose a new mapping architecture for StyleGAN to better disentangle a target attribute. This results in a learned blending between style codes, resulting in fine-grained local control of the edited images. They also introduce an additional latent code for controlling global aspects of the images (\eg{} pose, lighting, background). Finally, StyleMapGAN~\cite{kim2021stylemapgan} suggest an architectural change where the global \w{} latent code is replaced with a spatial map, and the global style infusion layer (\ie{} AdaIN or weight modulation) is replaced with a spatially adaptive one. This allows blending two images very naturally, with a high level of detail and finer local control (See Figure~\ref{fig:edits}f).

Studying the wide and versatile editing works, it is clear that latent-based editing holds great potential and sparks the curiosity of many. Some of the most recent works present unprecedented quality, showcasing the expressive powers of GANs in general and of StyleGAN in particular. However, all of these works still operate in lab conditions. They present a handful of novel editing operations. These, however, are still restricted (\eg{} only specific expressions can be altered and the degree of possible changes in pose is limited). These restrictions pose practical challenges when employing StyleGAN for industrial or in-the-wild use. Furthermore, they bear the burden of the generator's limited capabilities regarding the versatility and structure of the training data (see Section~\ref{sec:architecture}). In the future, we will probably witness more works that adapt to new data on the fly, possibly using techniques such as fine-tuning (Section~\ref{sec:fine-tuning}), or layer mixing, where several different models are trained, and their layers are mixed according to specific applications \cite{pinkney2020resolution,park2020swapping}. In any case, it seems that a core challenge editing works face is the evaluation of their quality, as discussed in Section~\ref{sec:metrics}.

\section{Encoding and Inversion}
\label{sec:encoding}

\begin{figure}
\setlength{\tabcolsep}{1pt}
    \centering
    \includegraphics[width=0.965\columnwidth]{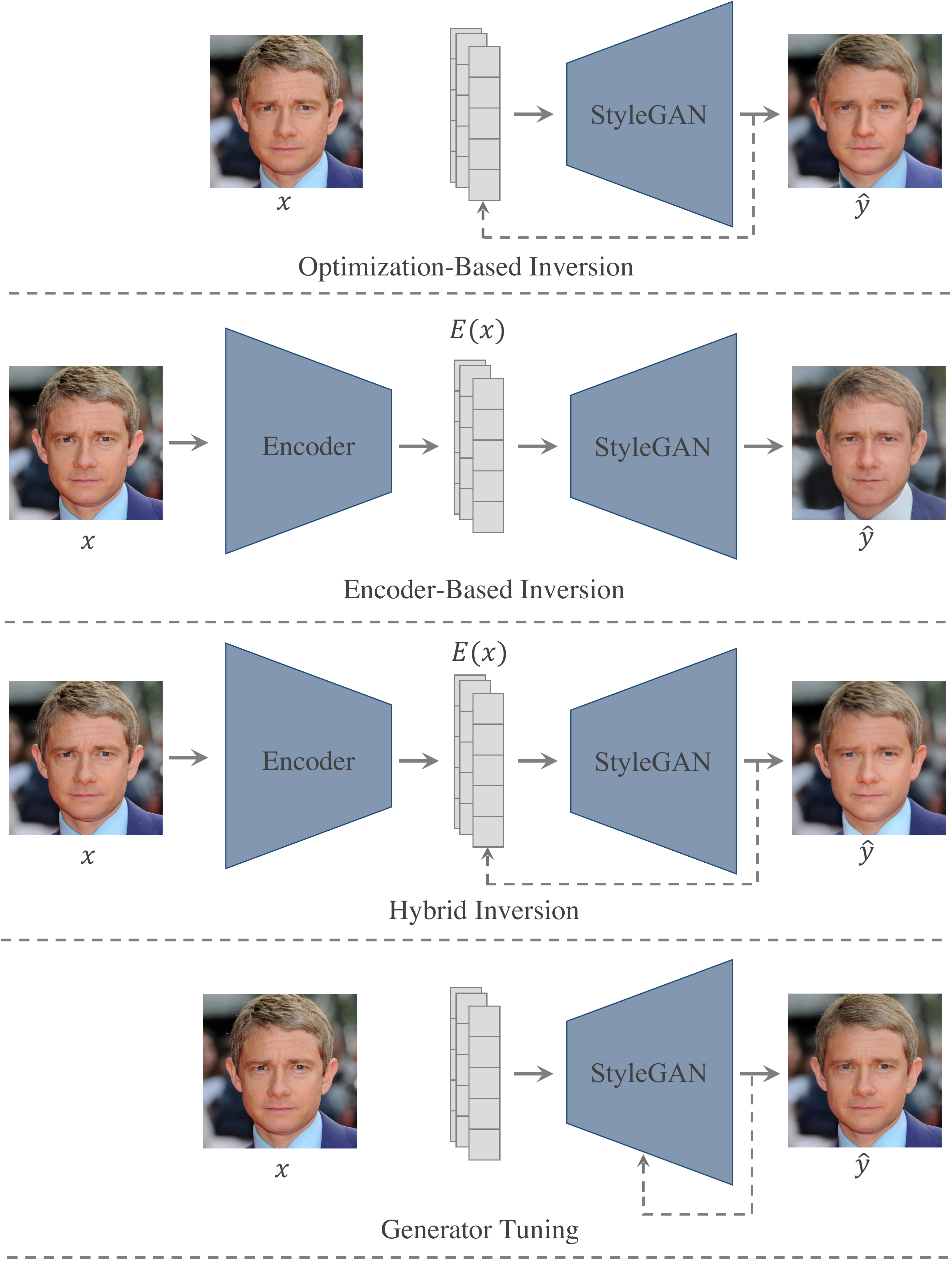}
    \caption{Various approaches for GAN Inversion. \textbf{Optimization-based} techniques perform a per-image optimization procedure on the latent vector to minimize the reconstruction loss between $x$ and $\hat{y}$. \textbf{Encoder-based} schemes aim to learn a direct mapping between the image $x$ to its latent representation $E(x)$. \textbf{Hybrid} techniques attempt to combine ``the best of both worlds`` by initializing the optimization procedure with the inversion prediction of a trained encoder. Finally, recent \textbf{generating tuning} methods fix a latent code and learn to modify the generator itself to obtain the reconstruction of the given image.
    Figure layout adopted from Xia~\etal~\shortcite{xia2021gan}.}
    \label{fig:inversion_schemes}
\end{figure}

The success of the aforementioned latent space editing techniques results in a natural question of how to apply such techniques to edit real images (i.e., images not necessarily residing within the GAN's domain). 
To do so, we need to find the latent representation of a given image, a task commonly referred to as \textit{GAN Inversion}. First introduced by Zhu~\etal~\shortcite{zhu2016generative}, the inversion task aims to find a latent vector from which a pre-trained GAN can most accurately reconstruct the given image. Formally, given an input image $x$, we want to minimize the distortion of the reconstructed image obtained from the inverted latent code $w$ using a well-trained generator $G$:
\begin{align}~\label{eq:opt}
w^* = \underset{w}{\arg\min} \mathcal{L} \left ( {x, G(w)} \right ),
\end{align}
where $\mathcal{L}$ is some reconstruction loss (e.g., the LPIPS perceptual loss~\cite{zhang2018unreasonable} and/or the pixel-wise L2 loss). In the following, we explore the various core approaches for performing this inversion process, outlined in Figure~\ref{fig:inversion_schemes}.

\subsection{GAN Inversion}~\label{sec:inversion}
Existing optimization-based GAN inversion methods search for the desired latent vector via a per-image latent vector optimization by solving Equation~\ref{eq:opt}~\cite{lipton2017precise,creswell2018inverting,abdal2019image2stylegan,abdal2020image2styleganpp,semantic2019bau,zhu2020improved,zhu2016generative,yeh2017semantic,gu2020image,wulff2020improving}.
Early works performing optimization attempted to invert into StyleGAN's learned latent space $\mathcal{W}$.
However, it has been shown that inverting a real image into a $512$-dimensional vector $w\in \mathcal{W}$ is not expressive enough to accurately encode and reconstruct real images. As such, it has become common practice to invert images into an extended latent space $\mathcal{W}+$~\cite{abdal2020image2styleganpp} defined by a concatenation of multiple $w$ vectors, one for each input of StyleGAN. 
While optimization techniques often result in near-perfect reconstructions of the input, they typically require several minutes to do so for a single image. 

To accelerate this optimization process, some works trained an encoder over a large collection of images to learn a direct mapping from an image to its latent representation~\cite{perarnau2016invertible,luo2017learning}. Here, the training objective can be defined by,
\begin{align}
\theta_E^* = \underset{\theta_{E}}{\arg\min} \sum_{i} \mathcal{L} ( \textbf{x}_i, G( E_{\theta_E}(\textbf{x}_i) ) ),
\end{align}
where the weights $\theta_E^*$ of the encoder are sought. 
Pidhorskyi~\etal~\shortcite{pidhorskyi2020adversarial} propose a StyleGAN-based autoencoder, where the encoder network $E$ is trained alongside the generator.

Many works have explored various avenues for improving the performance of encoder-based inversion methods in an attempt to close the gap in performance with optimization techniques.
Some have explored various encoder architectures for improving the inversion quality. 

Richardson~\etal~\shortcite{richardson2020encoding} and Xu~\etal~\shortcite{xu2021generative} explore a hierarchical encoder based on a feature pyramid network (FPN) to better match the coarse, medium, and fine-level details of StyleGAN's hierarchical structure. 
For extracting the learned styles from the encoder's feature maps, Richardson~\etal~\shortcite{richardson2020encoding} introduce $18$ separate map2style modules, one for each input layer of StyleGAN.
Wei~\etal~\shortcite{wei2021simple} and Alaluf~\etal~\shortcite{alaluf2021restyle} find that a complex hierarchical encoder is unnecessary, especially in unstructured domains (e.g., cars, churches, horses) and instead propose simpler backbones. Wei~\etal~\shortcite{wei2021simple} further replace the $18$ map2style blocks with a simple block comprised of a single average pooling layer and fully connected layer. 
Rather than encoding an image into a set of \textit{style vectors}, Kim~\etal~\shortcite{kim2021stylemapgan} instead invert images into an intermediate latent space with a spatial dimension, resulting in more accurate reconstructions compared to other encoder networks. They also demonstrate that this extended latent space enables reference-guided local edits of real images. More recently, Wang~\etal~\shortcite{wang2021high} explored inverting into multiple latent spaces to achieve higher-fidelity inversions. They first invert an image into $\mathcal{W}$, to capture low-frequency details. A second encoder is then trained to map the distortion map --- the difference between the given image and its initial inversion --- into a set of spatial feature modulation maps that capture the remaining high-frequency image information.

\setlength{\tabcolsep}{1pt}

\begin{figure}

{\small
\begin{tabular}{lccccc}

 \rotatebox[origin=t]{90}{Optimization} & 
  \raisebox{-.4\totalheight}{\includegraphics[width=0.185\columnwidth]{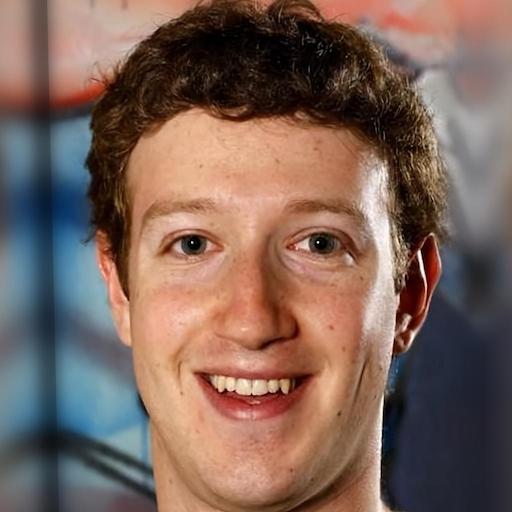}} & 
  \raisebox{-.4\totalheight}{\includegraphics[width=0.185\columnwidth]{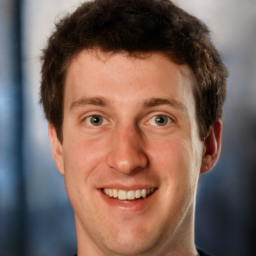}} & 
  \raisebox{-.4\totalheight}{\includegraphics[width=0.185\columnwidth]{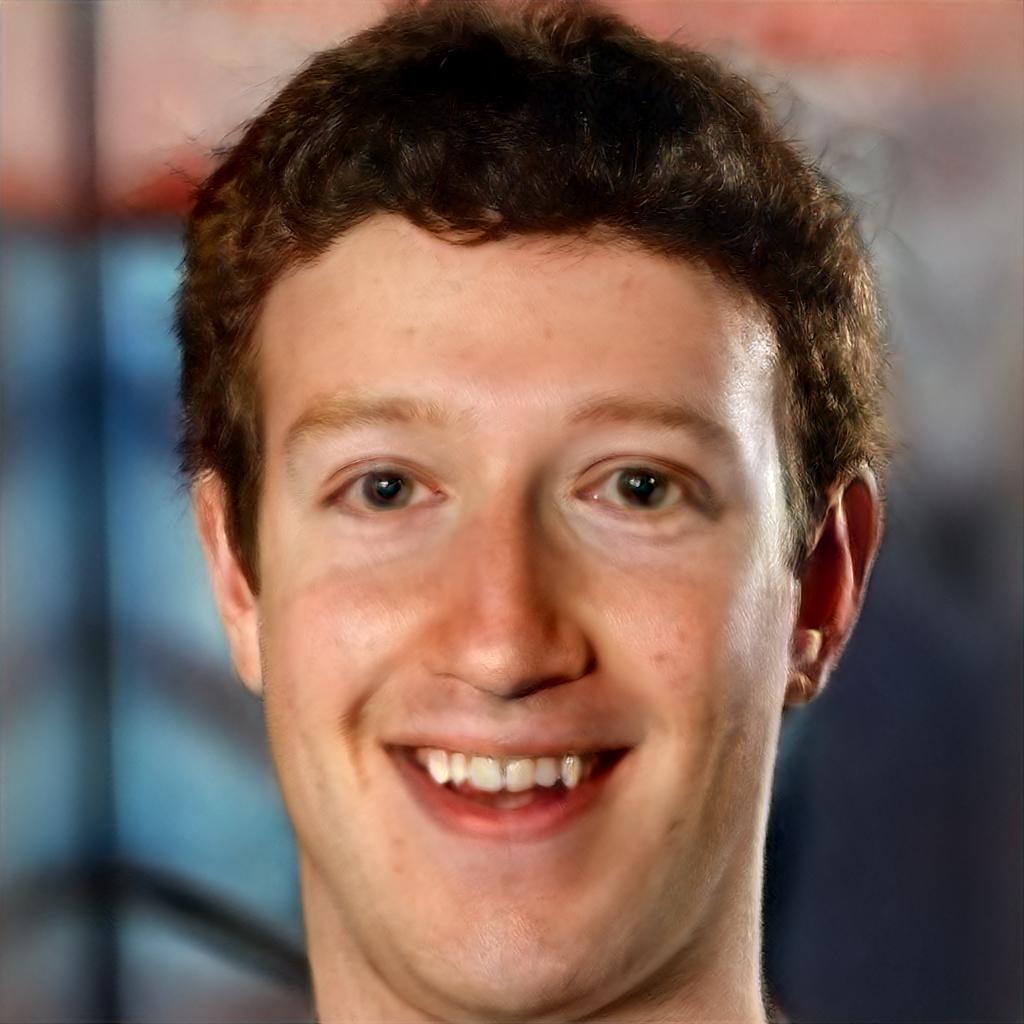}} & 
  \raisebox{-.4\totalheight}{\includegraphics[width=0.185\columnwidth]{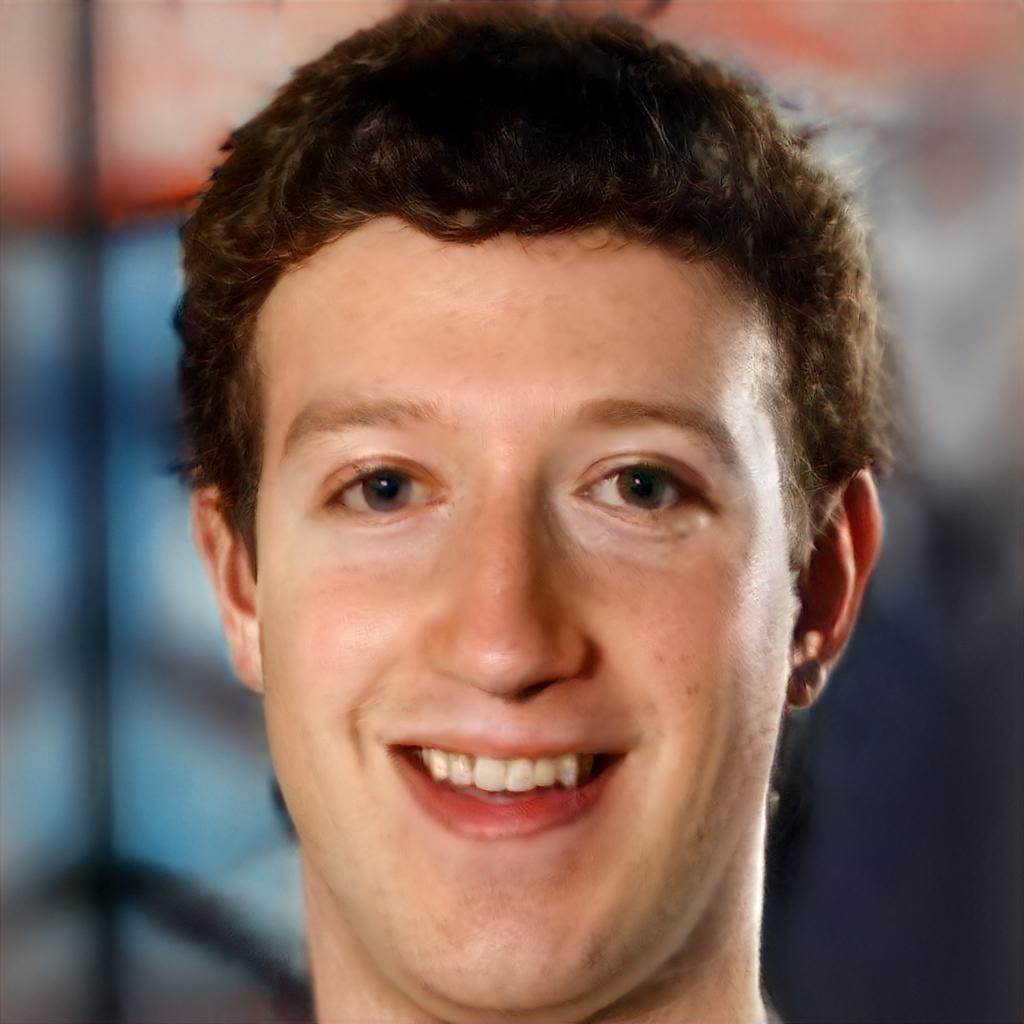}} & 
  \raisebox{-.4\totalheight}{\includegraphics[width=0.185\columnwidth]{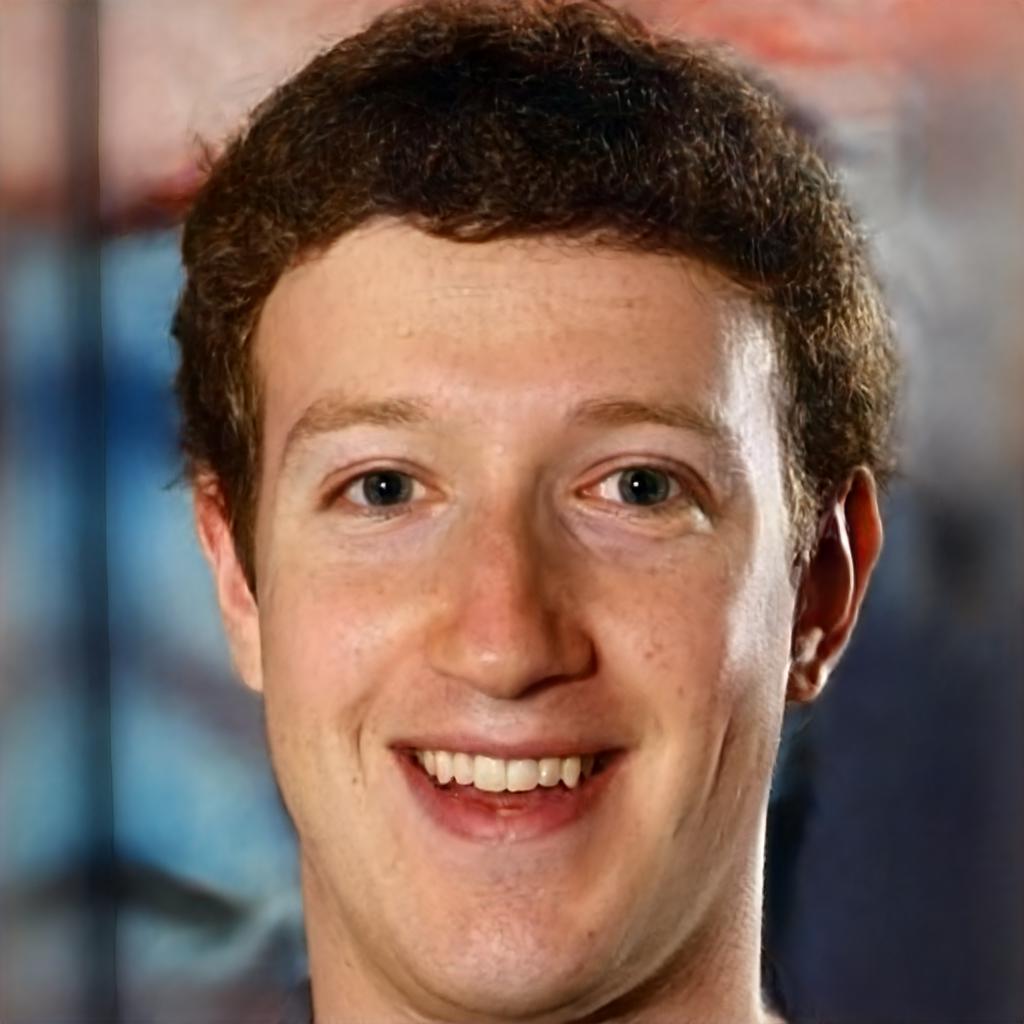}} \\
  \noalign{\vskip 1mm}
   & Input & Optim. $\mathcal{W}$ & Optim. $\mathcal{W}+$ & Hybrid & PTI \\
   \rotatebox[origin=t]{90}{Encoder} & 
  \raisebox{-.4\totalheight}{\includegraphics[width=0.185\columnwidth]{Figures/inversion/1721_input.jpg}} & 
  \raisebox{-.4\totalheight}{\includegraphics[width=0.185\columnwidth]{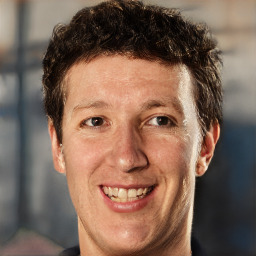}} & 
  \raisebox{-.4\totalheight}{\includegraphics[width=0.185\columnwidth]{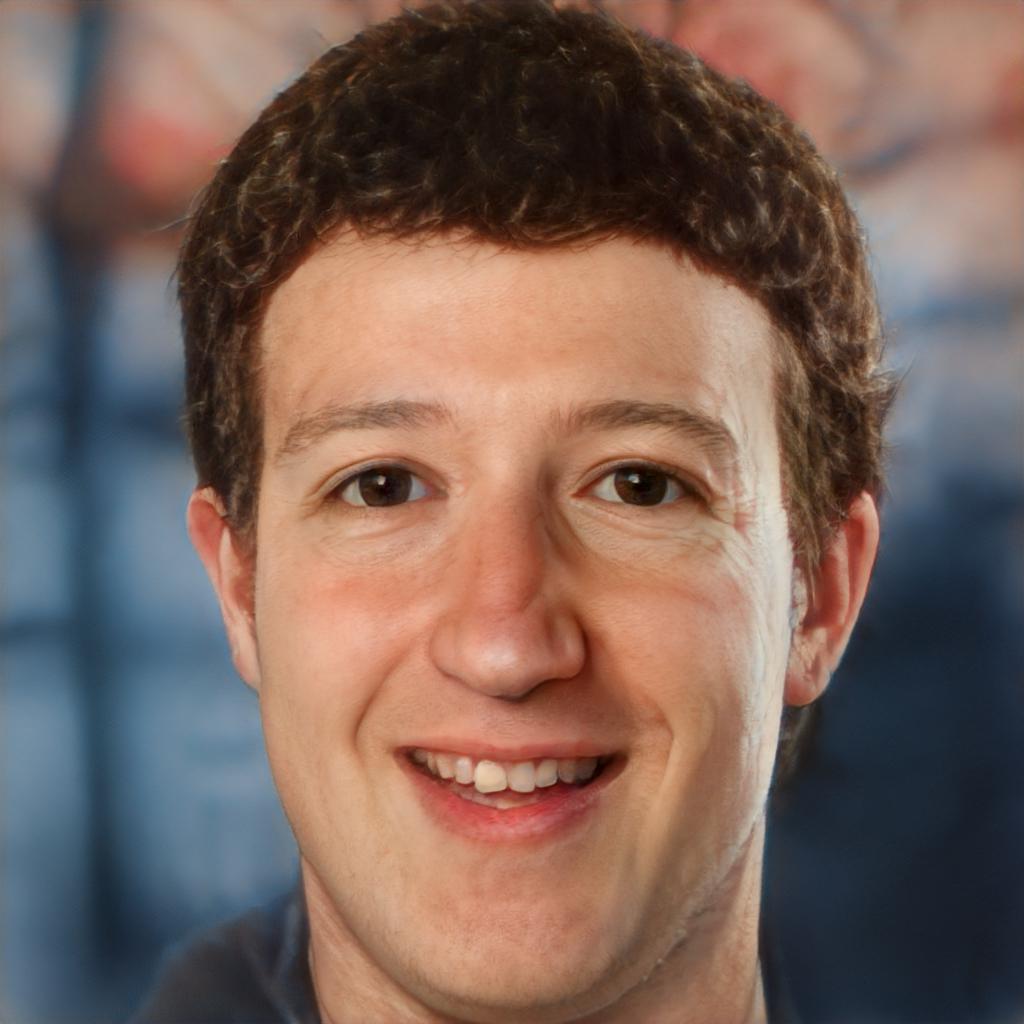}} & 
  \raisebox{-.4\totalheight}{\includegraphics[width=0.185\columnwidth]{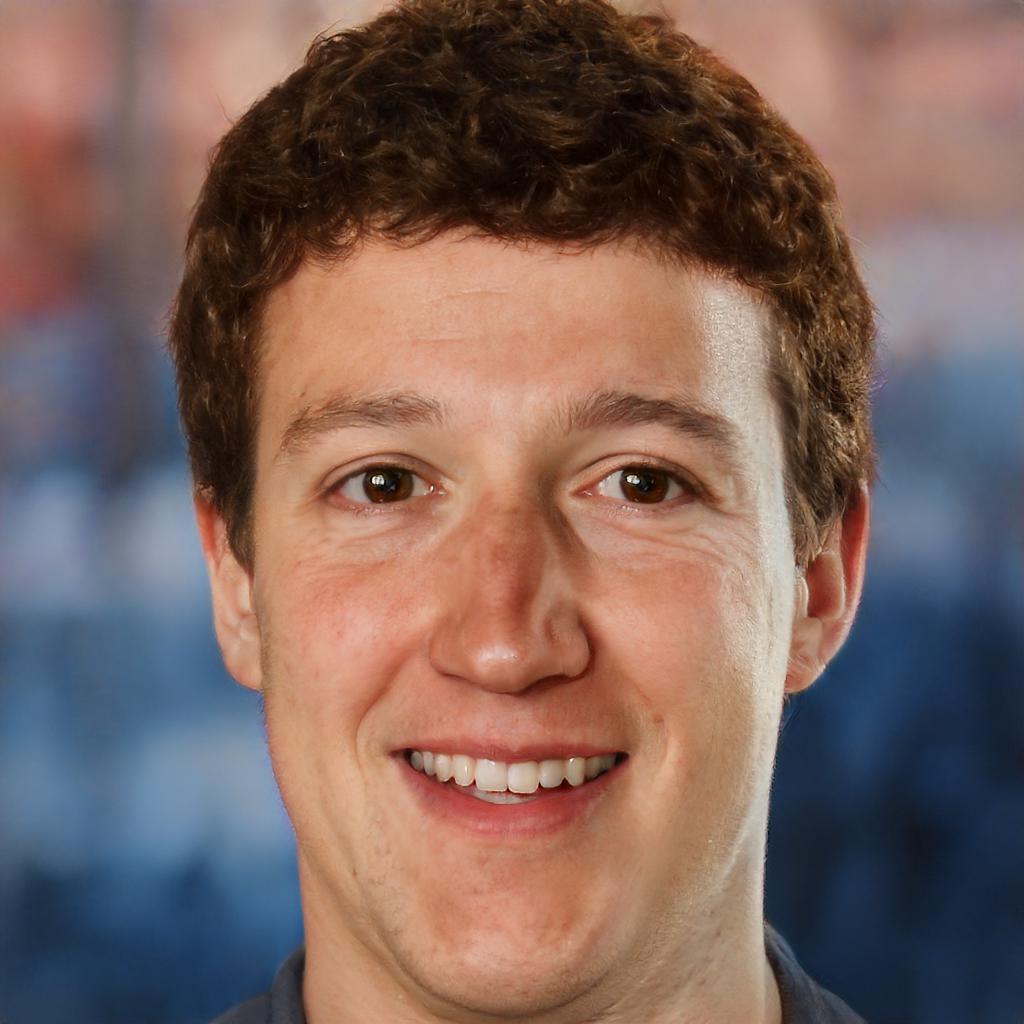}} & 
  \raisebox{-.4\totalheight}{\includegraphics[width=0.185\columnwidth]{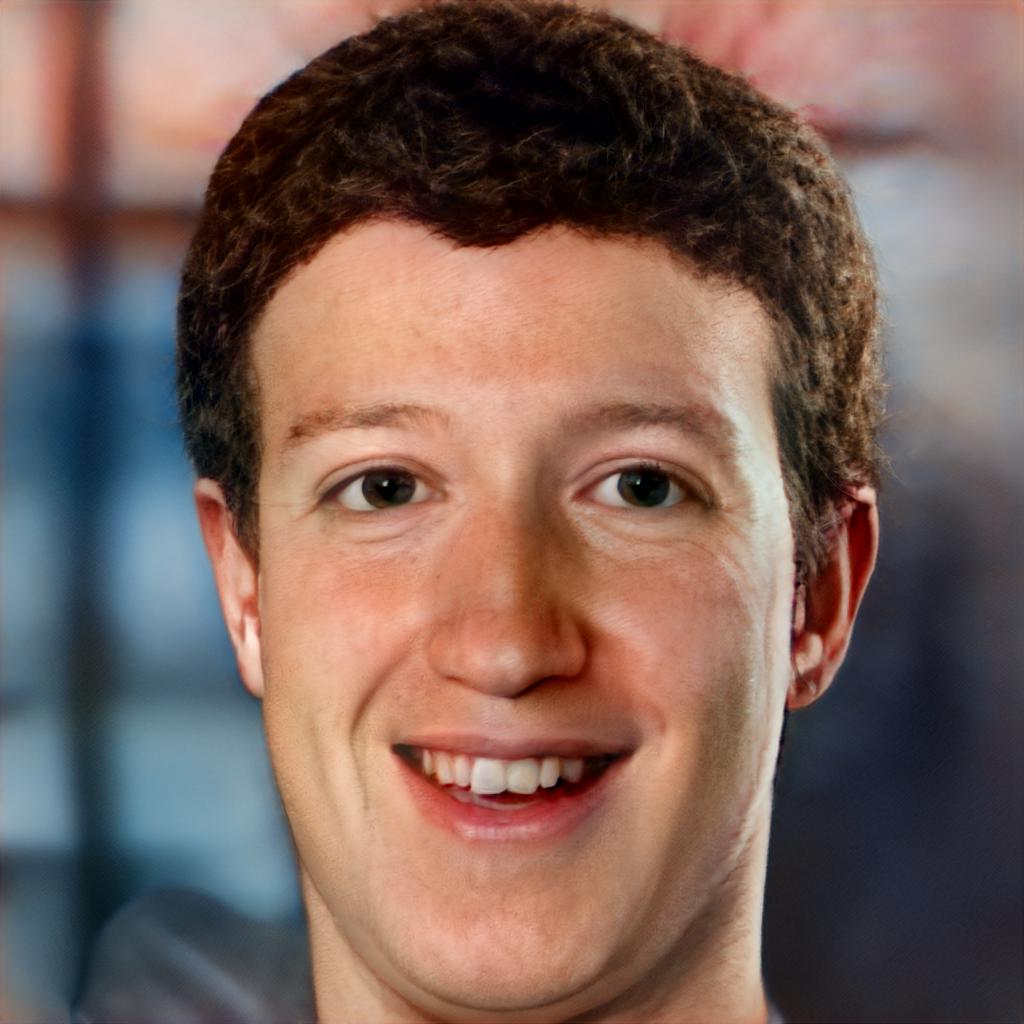}} \\
  \noalign{\vskip 1mm}
     & Input & IDInvert & pSp & e$4$e & ReStyle \\
   \rotatebox[origin=t]{90}{Editability} & 
  \raisebox{-.4\totalheight}{\includegraphics[width=0.185\columnwidth]{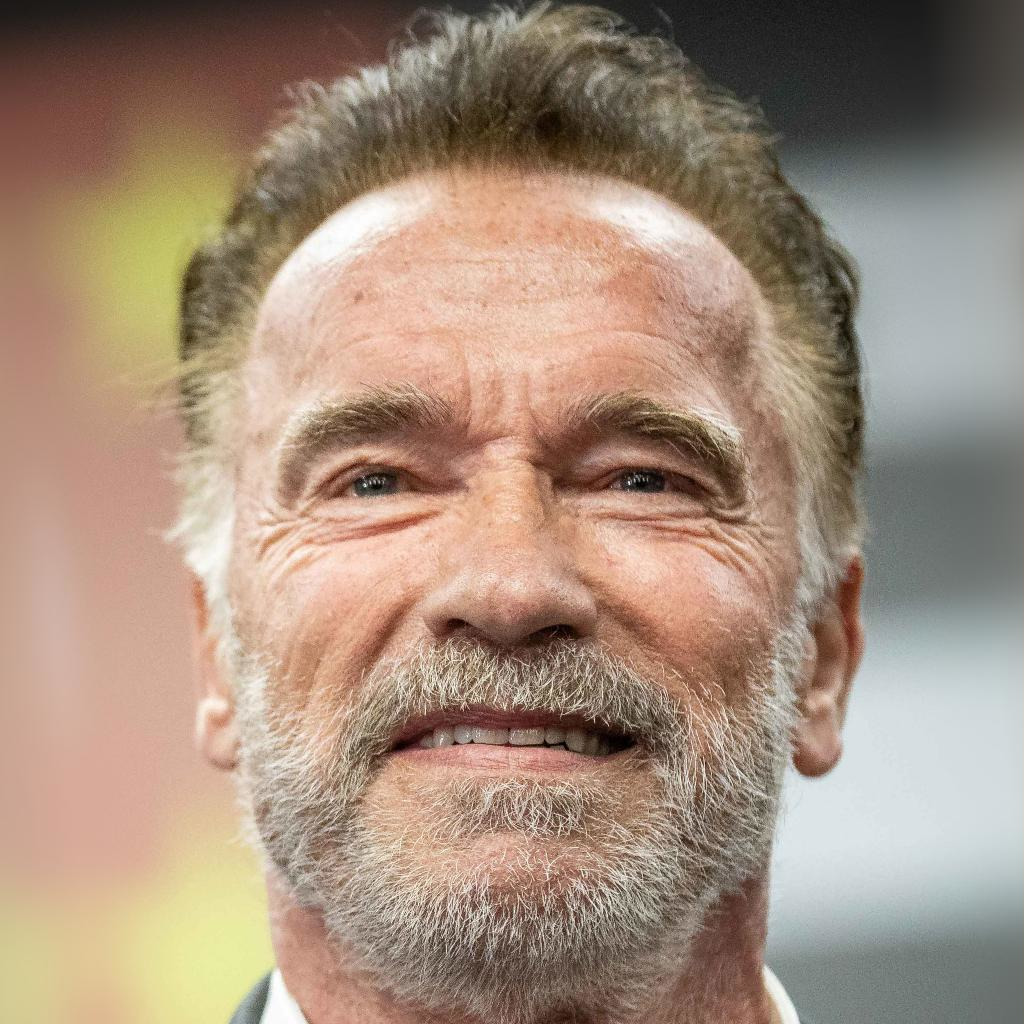}} & 
  \raisebox{-.4\totalheight}{\includegraphics[width=0.185\columnwidth]{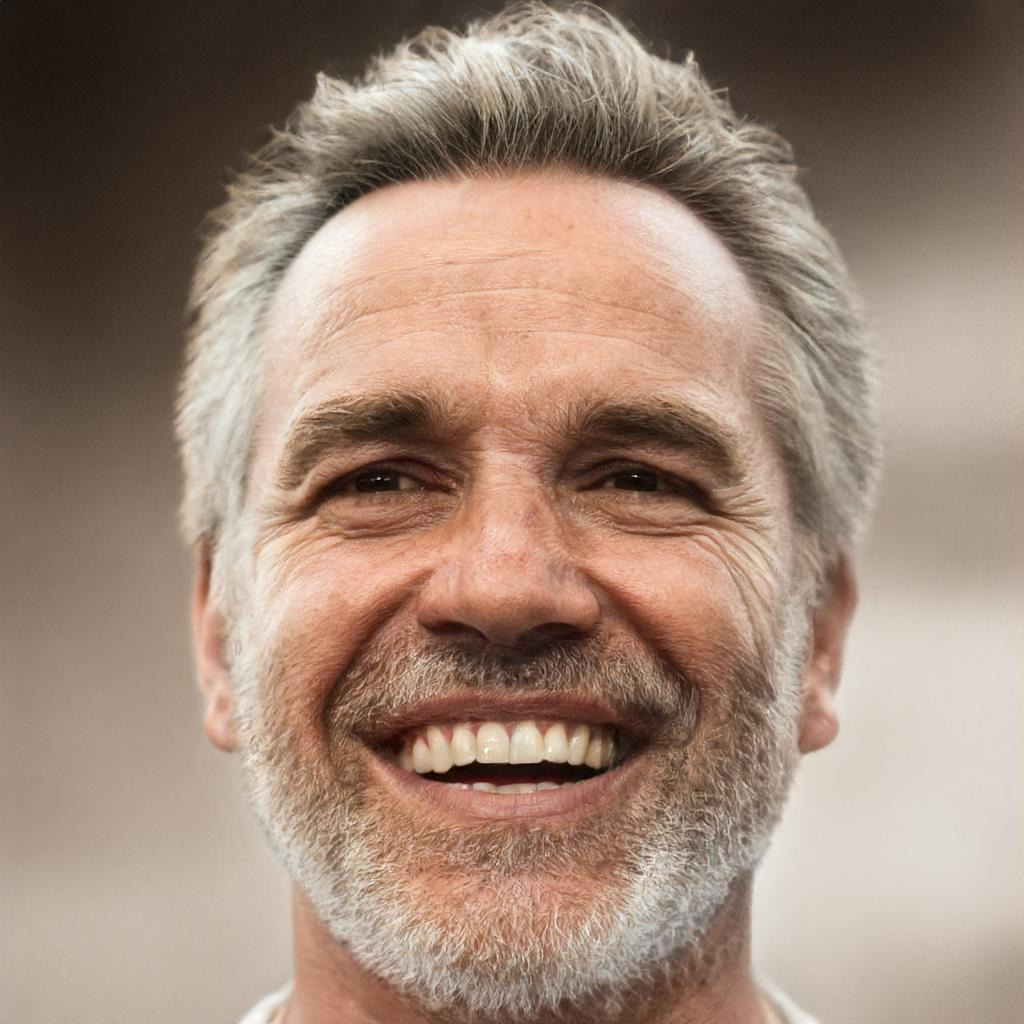}} &
  \raisebox{-.4\totalheight}{\includegraphics[width=0.185\columnwidth]{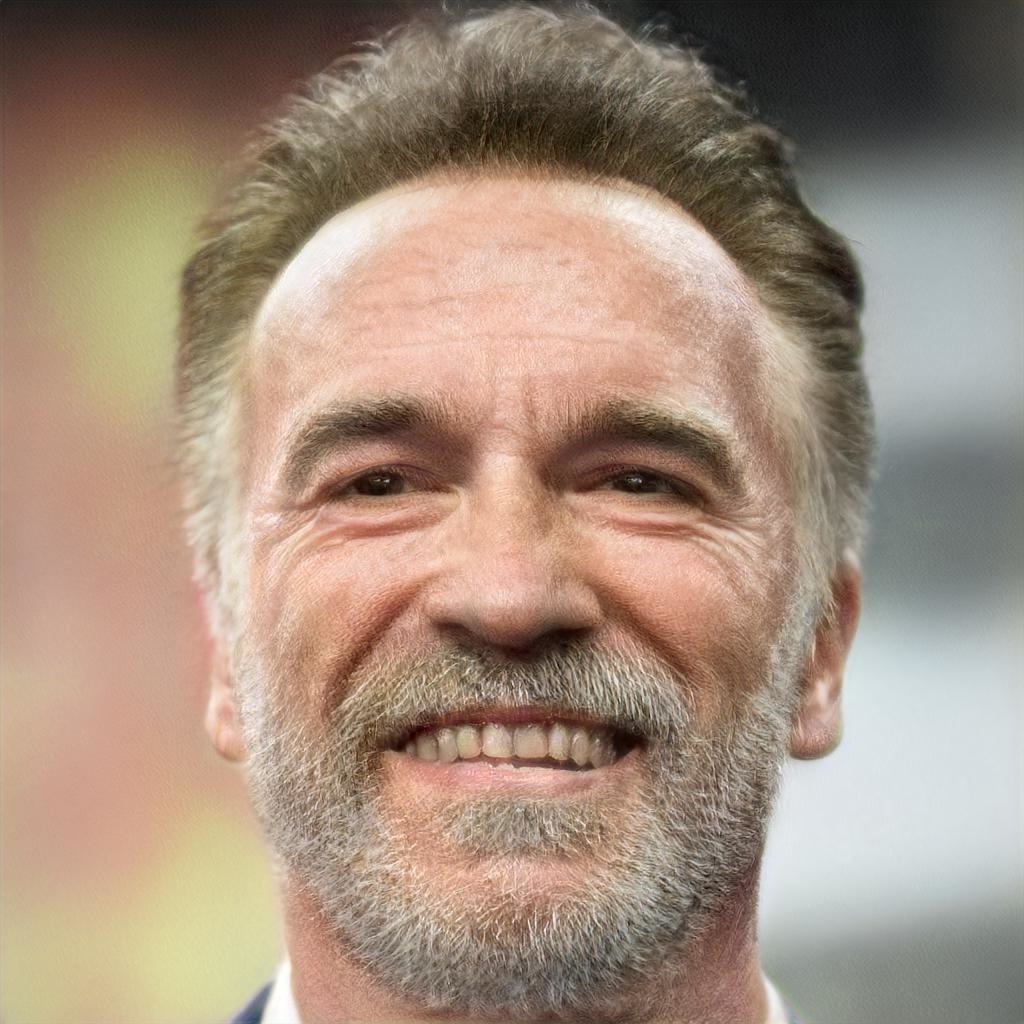}} & 
  \raisebox{-.4\totalheight}{\includegraphics[width=0.185\columnwidth]{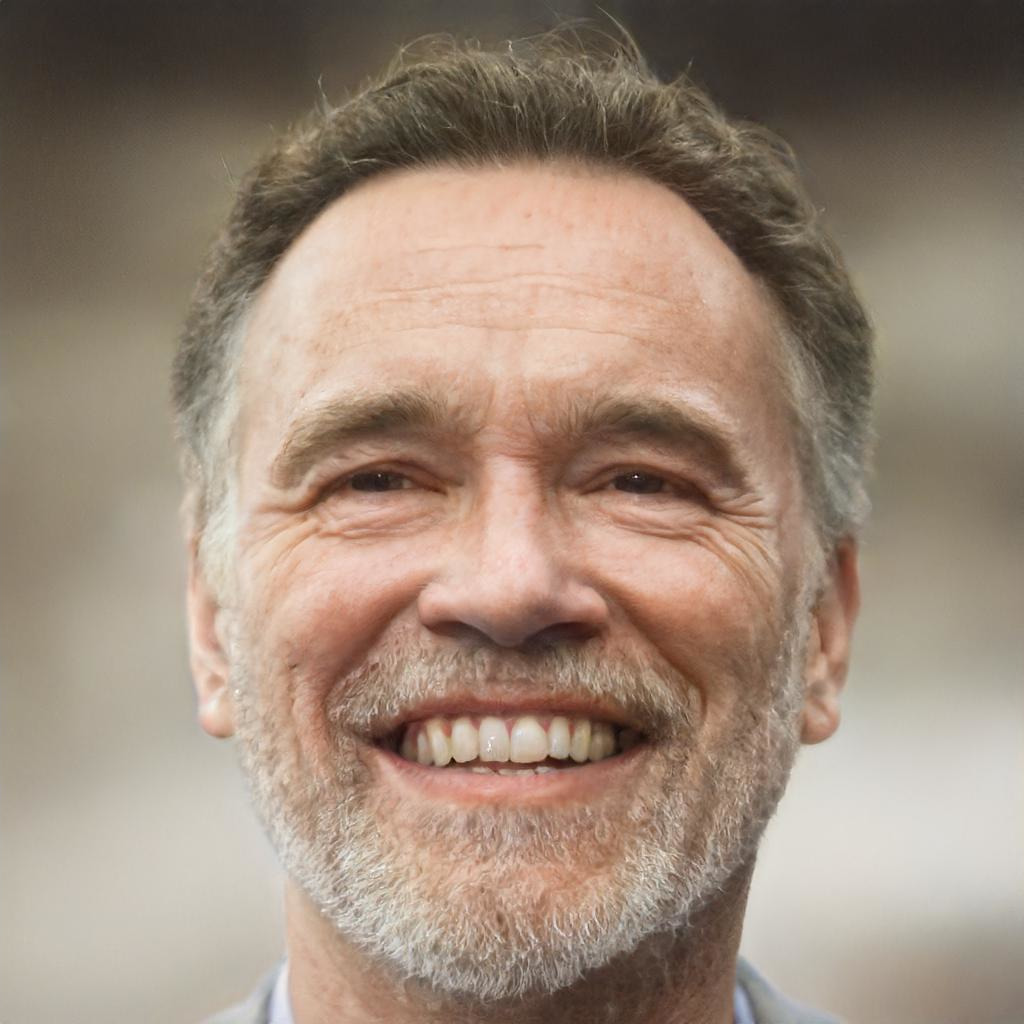}} & 
  \raisebox{-.4\totalheight}{\includegraphics[width=0.185\columnwidth]{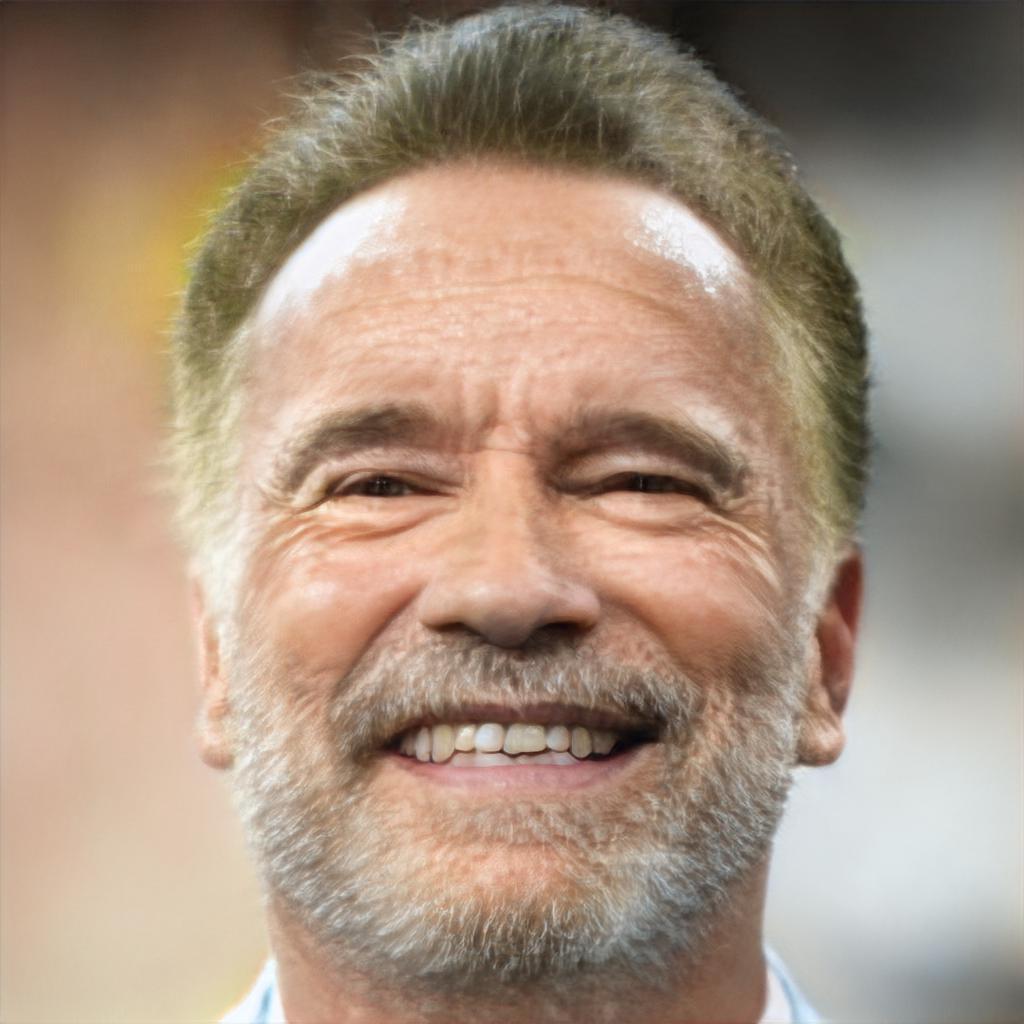}} \\ 
  \noalign{\vskip 1mm}
     & Input & Optim. $\mathcal{W}$ & Optim. $\mathcal{W}+$ & e$4$e & HyperStyle \\
   \rotatebox[origin=t]{90}{Out-Of-Domain} & 
  \raisebox{-.4\totalheight}{\includegraphics[width=0.185\columnwidth]{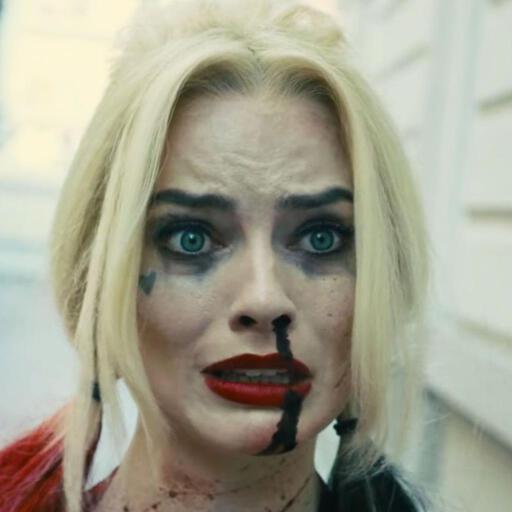}} & 
  \raisebox{-.4\totalheight}{\includegraphics[width=0.185\columnwidth]{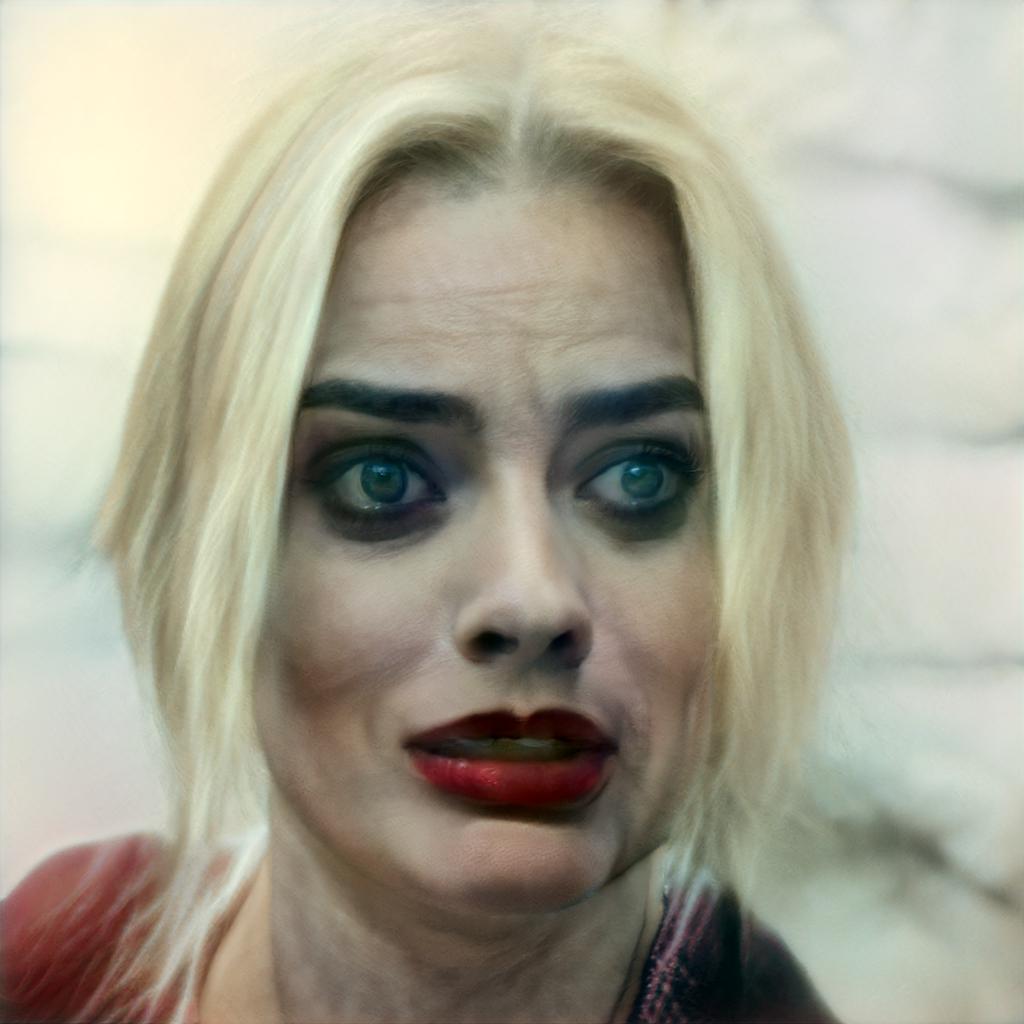}} & 
  \raisebox{-.4\totalheight}{\includegraphics[width=0.185\columnwidth]{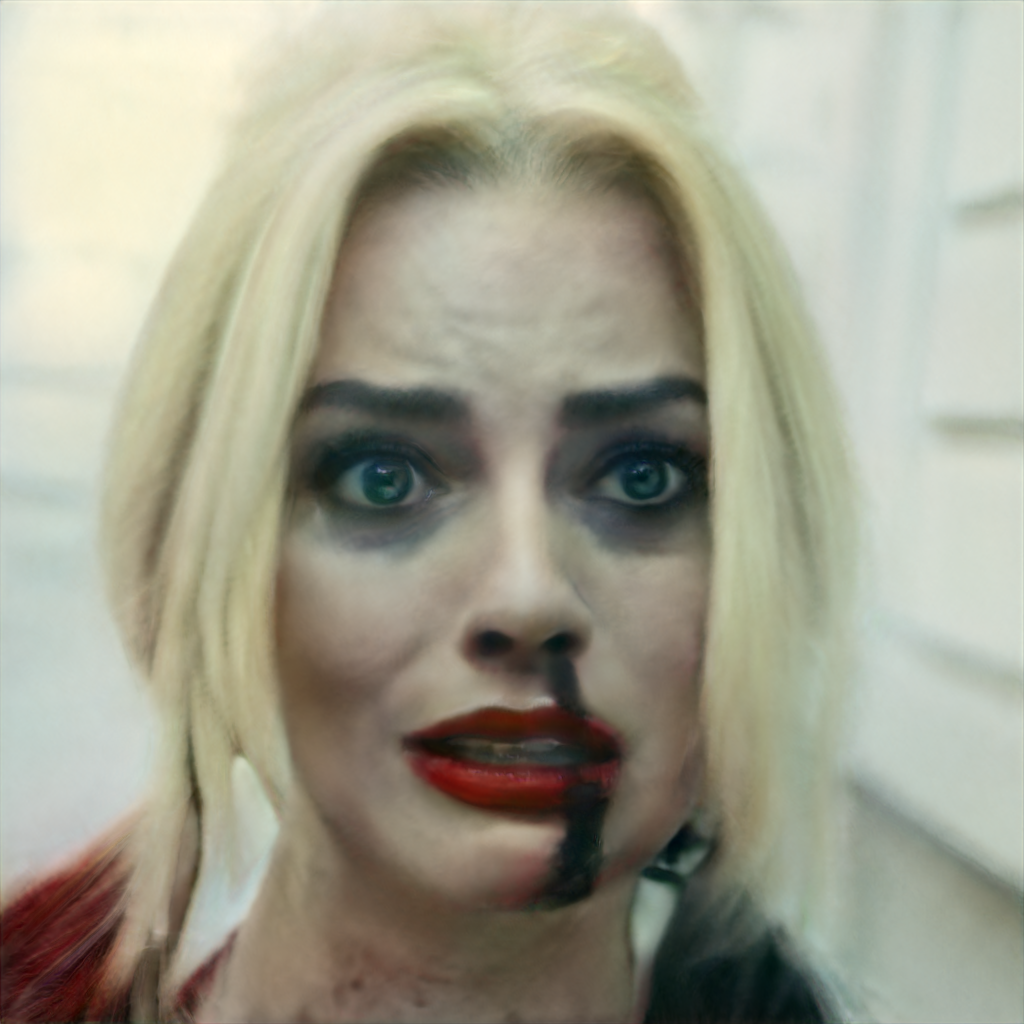}} & 
  \raisebox{-.4\totalheight}{\includegraphics[width=0.185\columnwidth]{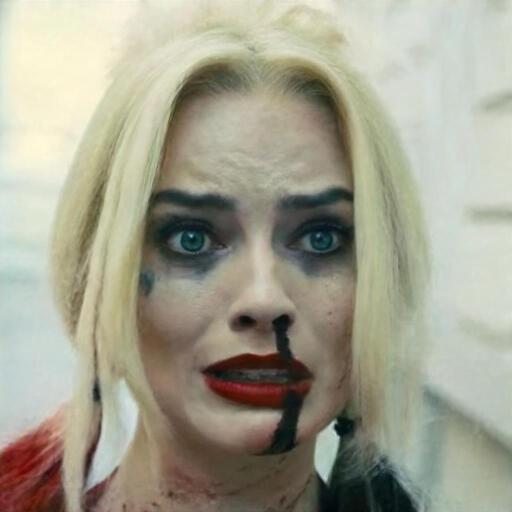}} &
   \raisebox{-.4\totalheight}{\includegraphics[width=0.185\columnwidth]{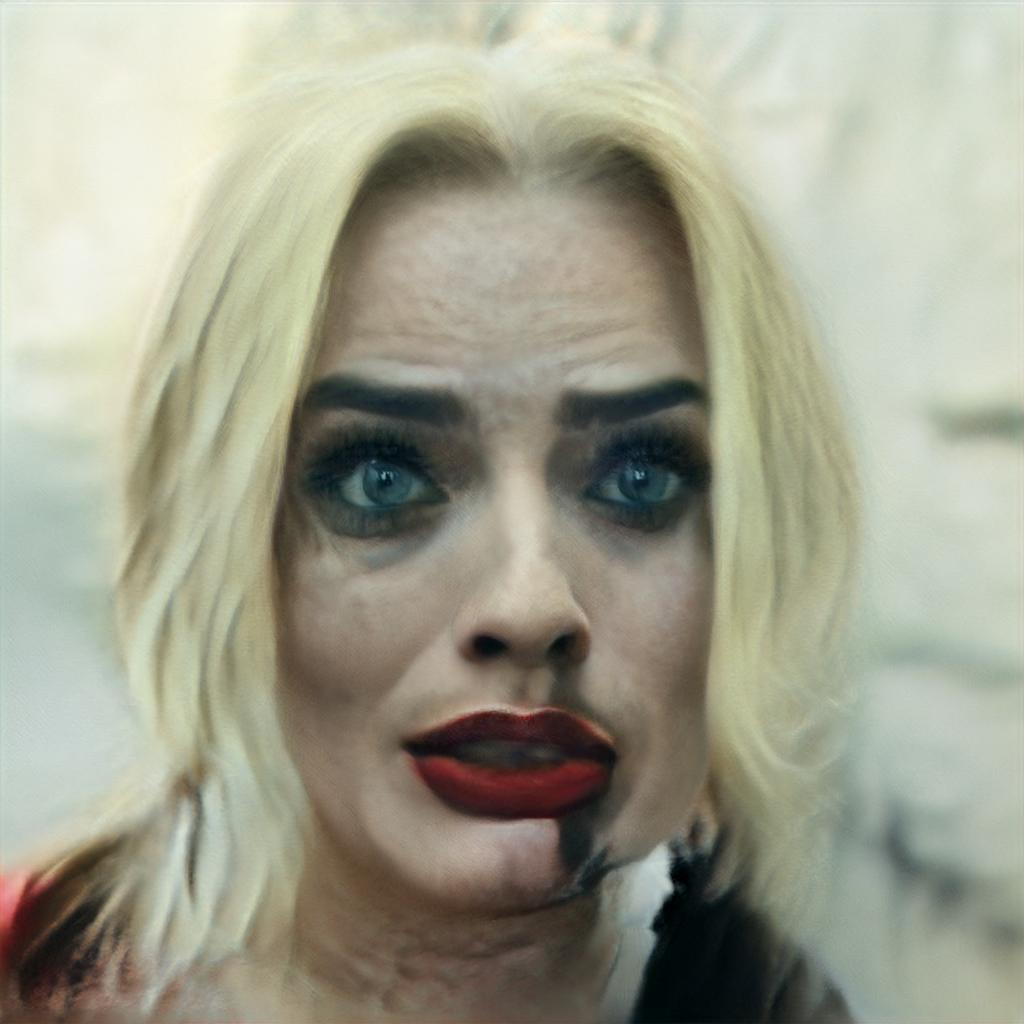}} \\
     & Input & ReStyle & Optim. $\mathcal{W}+$ & PTI & HyperStyle \\

\end{tabular}
}

\vspace{-0.1cm}
\caption{StyleGAN inversion. Upper row presents inversion results of optimization methods: optimization to $\mathcal{W}+$ as proposed by Karras \etal~\shortcite{karras2019analyzing}, Hybrid approach where pSp ~\cite{richardson2020encoding} and optimization to $\mathcal{W}+$ are employed, and PTI~\cite{roich2021pivotal}. Second raw demonstrates the inversion using encoders, IDInvert ~\cite{zhu2020domain}, pSp ~\cite{richardson2020encoding}, e$4$e ~\cite{tov2021designing} and ReStyle ~\cite{alaluf2021restyle}, over the same input image. The third row illustrates the editability of different regions in the latent space. The same smile editing was applied over inversion to $\mathcal{W}+$ space, $\mathcal{W}$ space and well-behaved regions of $\mathcal{W}+$ using the e$4$e ~\cite{tov2021designing} encoder. As can be seen, optimization to $\mathcal{W}+$ achieves high-quality reconstruction but poor editability. PTI mitigates this tradeoff by using $\mathcal{W}$ space and tuning the generator weights, but suffer from extensive time consumption. Like PTI, HyperStyle~\cite{alaluf2021hyperstyle} uses the $\mathcal{W}$ space for editing, but efficiently learn to modify the generator weights rather than perform time-intensive optimization. Lastly, the ability of PTI and HyperStyle to handle out-of-domain attributes, such as face painting, is presented at the bottom row. Zoom-in is recommended.} 
\label{fig:inv}
\vspace{-7pt}
\end{figure}

Another direction for improving the inversion of encoders is the improvement of the loss objectives used to learn the direct mapping from an image to its latent representation.
Zhu~\etal~\shortcite{zhu2020domain} employ a discriminator for an adversarial-based training of the encoder network and use the discriminator as an additional loss to the encoder.
To improve the inversion on the human facial domain, Richardson~\etal~\shortcite{richardson2020encoding} introduce a dedicated identity loss using a pre-trained facial recognition network. Tov~\etal~\shortcite{tov2021designing} extend this to additional domains by employing a similarity loss based on a MoCo~\cite{chen2020improved} feature extractor pre-trained on ImageNet. Wei~\etal~\shortcite{wei2021simple} utilize a pre-trained face parsing network to achieve more localized supervision during the encoder training.    

While encoder-based techniques result in an efficient inference scheme, taking a fraction of a second per image, the reconstructions are typically less accurate than optimization-based approaches.
In an attempt to close the gap between the two methodologies, Alaluf~\etal~\shortcite{alaluf2021restyle} introduce an iterative refinement scheme over standard encoder-based inversion techniques. Instead of directly outputting the inferred latent code using a single forward pass through the network, the encoder outputs a sequence of residuals used to iteratively improve the inverted latent code and corresponding reconstruction. Others~\cite{zhu2016generative,zhu2020domain} exploit the advantages of both of the above approaches and employ a hybrid technique combining the two. First, an initial approximate latent code $w_{initial}$ is inferred via a trained encoder. This latent code is then used to initialize the optimization procedure. In~\cite{guan2020collaborative}, the encoder network is used to initialize an optimization process, which in turn supervises the training of the encoder network via a set of reconstruction losses. 
We refer the reader to Figure~\ref{fig:inv} for a comparison of various optimization-based and encoder-based inversion techniques. 
Xia \etal~\shortcite{xia2021gan} provide a comprehensive survey and analysis of recent inversion methods, exploring the three aforementioned methodologies and their use in various editing applications. 

While the inversion process is a well-studied problem, it remains an open challenge. Numerous works~\cite{abdal2020image2styleganpp,zhu2020improved,tov2021designing,zhu2020domain,wulff2020improving} demonstrate the existence of a reconstruction-editability trade-off. Whereas $\mathcal{W}+$ has been shown to be more expressive than $\mathcal{W}$~\cite{abdal2020image2styleganpp}, supporting more accurate reconstructions, its use leads to latent codes which lie in regions of the latent space that were unobserved during the generator training. In these regions, the semantic structure of the latent space deteriorates, resulting in degraded performance of latent space traversal editing methods, as demonstrated in Figure.~\ref{fig:inv}.
Some works searched for a good point on this trade-off curve.
Tov~\etal~\shortcite{tov2021designing} design an encoder to embed images into $\mathcal{W}+$ that are close to $\mathcal{W}$, resulting in a good balance between reconstruction quality and editability. Zhu~\etal~\shortcite{zhu2020improved} analyze various latent spaces to achieve more control over the reconstruction-editability trade-off.
In an attempt to side-step this trade-off, Roich~\etal~\shortcite{roich2021pivotal} propose a pivotal tuning method to inject new identities into well-behaved, editable regions of StyleGAN's latent space. They first use a standard optimization procedure to find a latent code $w\in\mathcal{W}$ approximating the input image. This is followed by a per-image fine-tuning session where the generator weights are modified to improve the reconstruction quality. Other generator tuning approaches have also been proposed for achieving high fidelity reconstructions (see Section~\ref{sec:fine-tuning}).

While most generator tuning approaches improve the image inversion via a per-image optimization of the generator weights, such an approach is costly in terms of inference time. To reduce this inference overhead, Alaluf~\etal~\shortcite{alaluf2021hyperstyle} and Dinh~\etal~\shortcite{dinh2021hyperinverter} propose a hypernetwork-based encoder that \textit{learns} how to modify the pre-trained generator weights to best reconstruct a given image. Such a learned approach results in high-fidelity reconstructions and edits, at a fraction of the time compared to optimization-based tuning approaches.

Finally, while most works studying inversion focus on encoding and editing still images, when it comes to video editing new challenges arise. Specifically, video inversion should be temporally consistent. Tzaban~\etal~\shortcite{tzaban2022stitch} demonstrate that by combining encoders~\cite{tov2021designing} with generator tuning techniques~\cite{roich2021pivotal}, the consistency of the original video can be maintained. Another challenge can be found in the texture-sticking phenomenon observed in StyleGAN1 and StyleGAN2~\cite{karras2021alias}, which hinders the realism of generated and manipulated videos. To overcome this, Alaluf~\etal~\shortcite{alaluf2022times} combine the PTI~\cite{roich2021pivotal} and ReStyle~\cite{alaluf2021restyle} encoding techniques for encoding and editing videos with the StyleGAN3~\cite{karras2021alias} generator. Further leveraging the equivariance of StyleGAN3, they demonstrate the ability to expand the field of view when working on a video with a cropped subject resulting in more uniform video editing. 

\subsection{Latent Space Embedding}
\label{sec:endcoding}

Image inversion provides a latent code that reconstructs a given image. As the image itself is given, the produced latent code is usually not of interest on its own. Rather, one applies inversion to then manipulate the latent code to produce a new latent code that corresponds to a novel image.

In this light, the limitations of inversion are clear. First, inversion methods require that the input image be invertible. That is, it must reside within one of the latent spaces of StyleGAN. Second, it is assumed that there is a known global transformation in the latent space to produce the desired manipulated code. However, for some applications, at least one of these assumptions is not true. For example, consider commonly studied image-to-image tasks such as semantic map-to-image and sketch-to-image \cite{isola2017image}. As StyleGAN is trained on one domain, usually the natural image domain, the sketch image would not be invertible to StyleGAN's latent space.

Hence, several works have analyzed this limitation and have proposed a broader task of \textit{Latent Space Embedding.} In this setting, for some image $x$, one seeks a function $f$ such that $G(f(x)) \sim h(x)$, where $G$ is the pretrained StyleGAN generator and $h$ is some conceptually known function in image space (e.g., sketch-to-image). Under this perspective, inversion is a special case in which $h$ is the identity function. However, many methods have proposed training such function $f$ for specific transformations $h$.

\begin{figure}
\setlength{\tabcolsep}{1pt}
    \centering
    \begin{tabular}{ c c }
        \includegraphics[width=0.45\columnwidth]{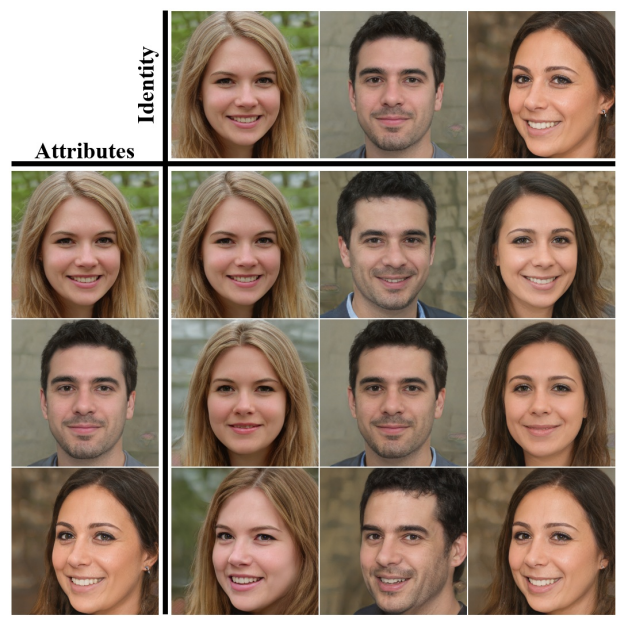} &  
        \includegraphics[width=0.45\columnwidth]{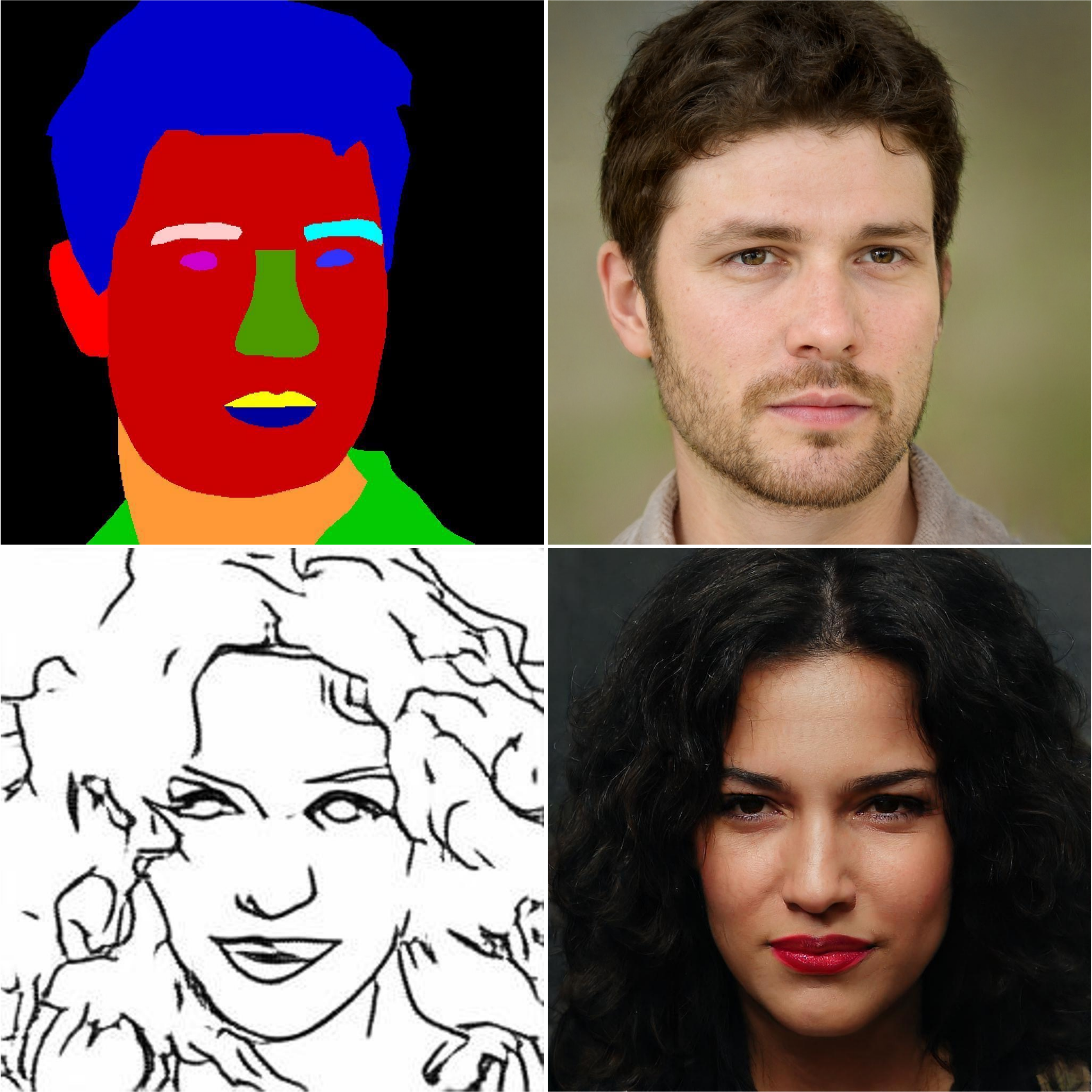} \\
        \rotatebox[origin=c]{0}{\shortstack{(a) ID disentanglement}} & 
        \rotatebox[origin=c]{0}{\shortstack{(b) pixel2style2pixel}} \\ %
        \includegraphics[width=0.45\columnwidth]{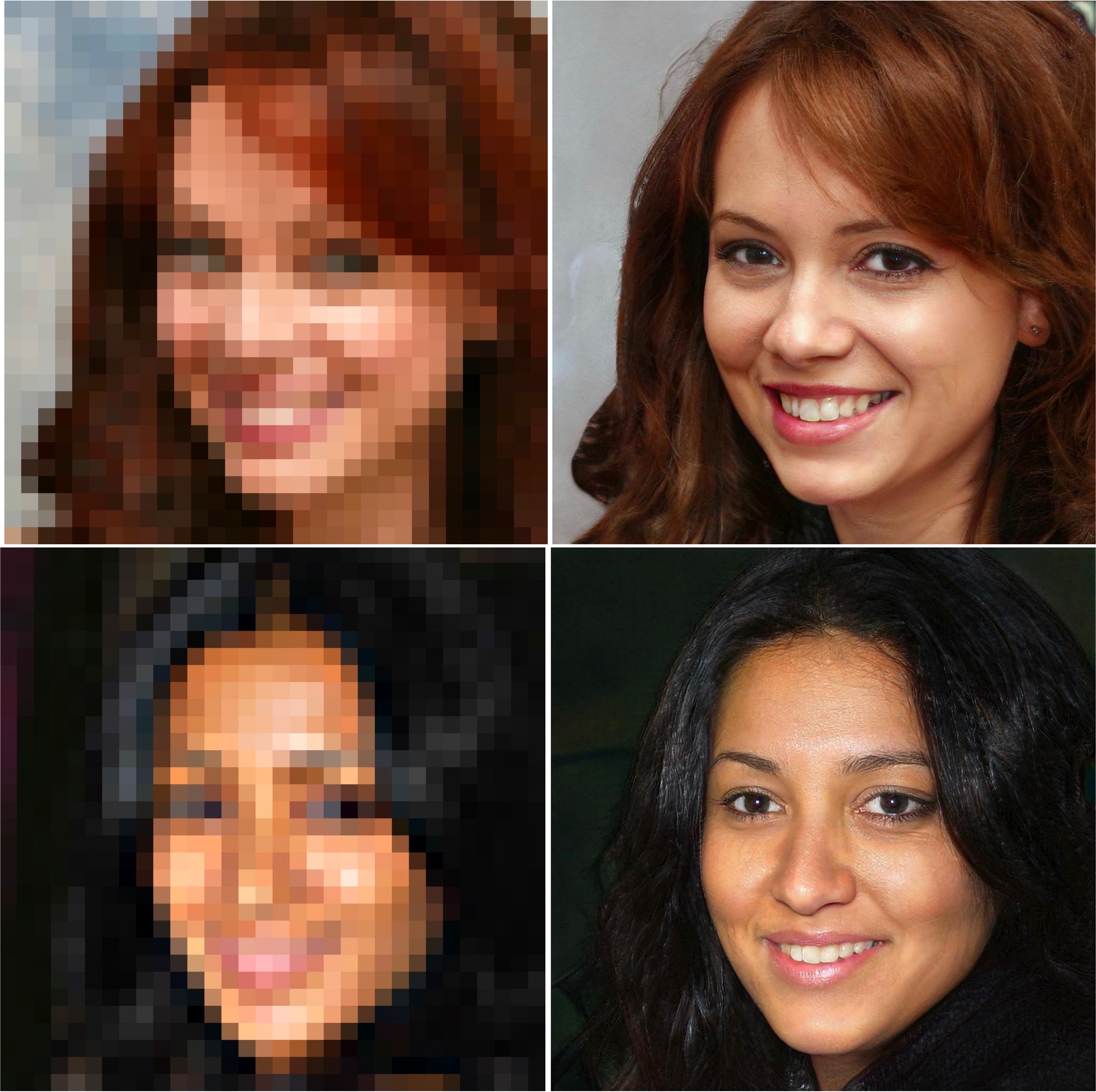} &
        \includegraphics[width=0.45\columnwidth]{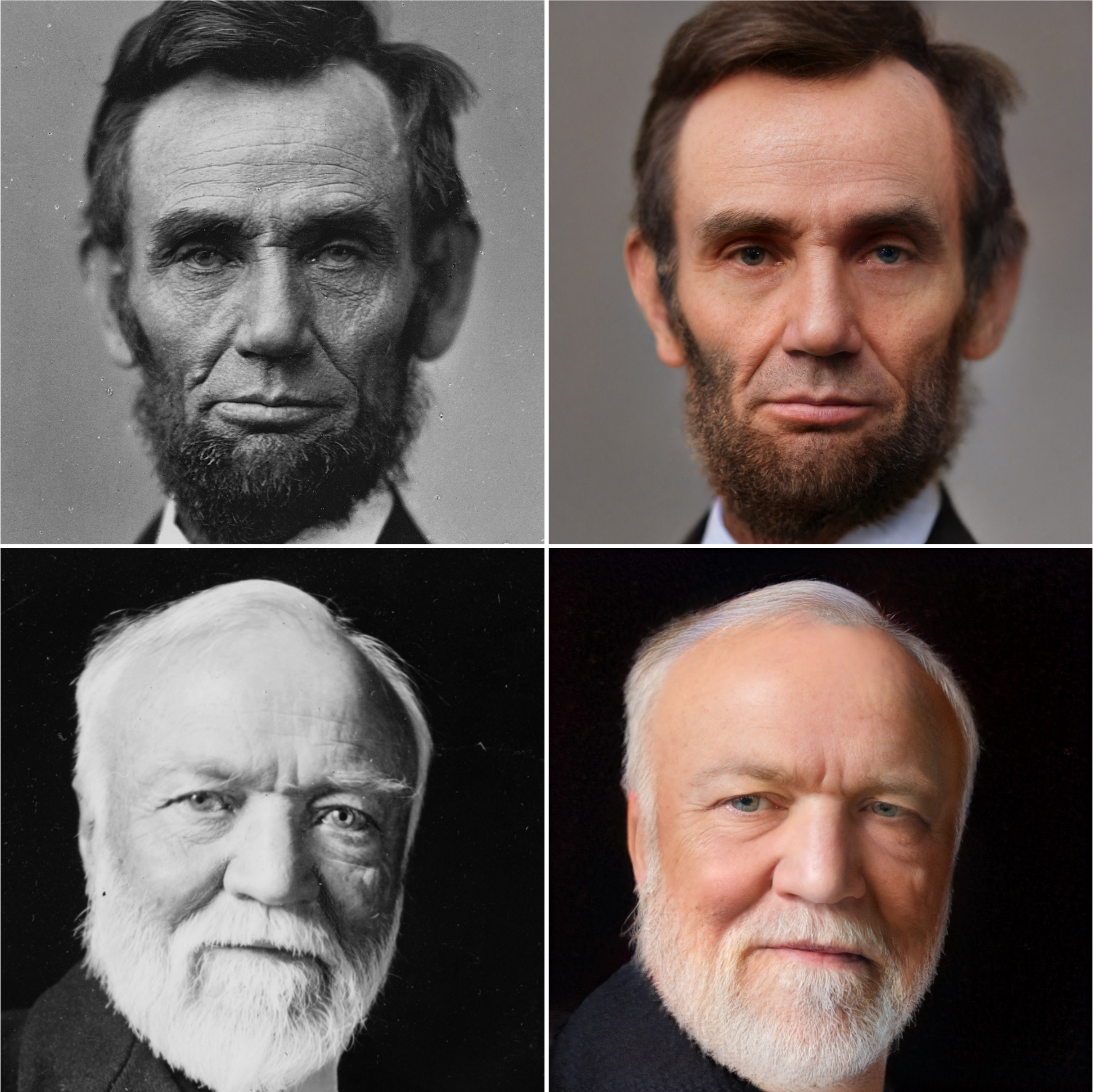} \\
        \rotatebox[origin=c]{0}{\shortstack{(c) PULSE}} & 
        \rotatebox[origin=c]{0}{\shortstack{(d) Time-Travel Rephotography}}  
    \end{tabular}
    \caption{Examples of prominent works leveraging latent space embedding. (a) Nitzan~\etal~\shortcite{Nitzan2020FaceID} disentangle identity from other face attributes and recompose them to generate novel face images. (b) pSp~\cite{richardson2020encoding} proposed a generic pix2pix-like architecture for embedding into StyleGAN's latent space. (c) PULSE ~\cite{menon2020pulse} perform super resolution by recovering the StyleGAN latent code that after downsampling reconstructs the original image.
    (d) Time-Travel Rephotography ~\cite{Luo-Rephotography-2020} restore old photos with a similar approach to PULSE, using a ``old photo" degredation module instead of downsampling.}
    \label{fig:latent_embedding}
\end{figure}

First, Nitzan~\etal~\shortcite{Nitzan2020FaceID} propose using StyleGAN to disentangle identity from other facial attributes and recompose novel images. They do so by extracting identity and attribute representations from different images, combining them, and then training a mapping network to directly produce the latent code that fuses the two representations, resulting in a novel face image, see Figure~\ref{fig:latent_embedding}(a).

Next, Richardson~\etal~\shortcite{richardson2020encoding} proposed a generic framework, pixel2style2pixel (pSp), to perform a wide variety of image-to-image tasks, such as the aforementioned sketch-to-face and semantic map-to-face. 
pSp employs an encoder architecture based on a feature pyramid network (FPN), to naturally match the StyleGAN hierarchical generative path. Through the right inductive bias, this work demonstrates state-of-the-art inversion quality, along with various other successful encoding tasks for human faces, including in-painting, super-resolution, unsupervised frontalization, colorization, and more, see Figure \ref{fig:latent_embedding}(b). 

Several works used the above concept of latent space embedding for a variety of tasks, often obtaining state-of-the-art performance. Most notably for the task of restoring corrupted images. PULSE \cite{menon2020pulse} solves super resolution of facial images. Specifically, PULSE performs a latent-space optimization to recover a code, from which StyleGAN synthesis followed by downsampling reconstructs the original low-resolution input image (see Figure \ref{fig:latent_embedding}(c)). Time-Travel Rephotography \cite{Luo-Rephotography-2020} (Figure \ref{fig:latent_embedding}(d)) restores old photographs, transforming them to modern imagery. They do so by following a similar approach to PULSE, with a dedicated degradation module replacing the down-sampling step. GFPGAN \cite{wang2021gfpgan} solves blind face restoration by constructing dedicated losses and architecture. GLEAN \cite{chan2021glean} use an encoder-latent bank-decoder architecture to solve super-resolution tasks. Once more, the decoder is a well-trained StyleGAN generator.

Of the aforementioned tasks, a task receiving considerable attention is that of sketch-to-image due to its immediate application to a variety of real-world settings. Building on the multi-modal sketch-to-image approach from pSp~\cite{richardson2020encoding}, Wei~\etal~\shortcite{wei2021simple} utilize a specialized face parsing loss to improve the alignment between an input sketch or semantic map and the output realistic facial image. Finally, Wang~\etal~\shortcite{wang2021sketch} modify a pre-trained StyleGAN for transforming a given sketch into a realistic image. As the generator tuning is subtle, the inherent characteristics of the original generator (e.g., color, texture) are well-preserved while supporting multi-modal synthesis. 

A myriad of other applications has also been explored via the task of latent space embedding. Chai~\etal~\shortcite{chai2021latent} train an encoder for performing image composition and image completion by leveraging the strong image prior of a pre-trained StyleGAN generator. Alaluf~\etal~\shortcite{alaluf2021matter} pair a pSp encoder~\cite{richardson2020encoding} and pre-trained age regressor~\cite{RotheDEX} for performing age transformation on real images via StyleGAN's latent domain. Jang~\etal~\shortcite{jang2021stylecarigan} transform real facial images to caricatures by altering the specific layers of a pre-trained StyleGAN. Specifically, they leverage the hierarchical nature of StyleGAN and modify the coarse input layers controlling head shape while keeping the fine layers controlling style and color unchanged. 

Xu~\etal~\shortcite{xu2021linear} and Ling~\etal~\shortcite{ling2021editgan}
edit a given image in the domain of its part-segmentation. The fundamental observation is that 
one can train a simple function, $f$ inferring a semantic segmentation, corresponding to the image generated by a latent code, from intermediate activations of StyleGAN on that latent code. Specifically, they use the up-sampled and concatenated per-layer activations as input to $f$. This observation and construction was concurrently proposed by other works \cite{tritrong2021repurposing, zhang2021datasetgan} for different applications and are discussed in Section \ref{sec:discriminative}.
Given an image, Xu~\etal~\shortcite{xu2021linear} propose to compute its semantic segmentation using off-the-shelf methods. Then, the segmentation map is edited with some desired effect and finally, it is embedded into the latent space of StyleGAN through StyleGAN's own layers as well as those of the function $f$. The resulting latent may then be forwarded through StyleGAN to generate the edited image.

Other works~\cite{wei2021simple,zhu2021shot,yang2021shapeediter} examined the task of face swapping by blending the latent representations of two input images embedded via a learned encoder.  Zhu~\etal~\shortcite{zhu2021barbershop} study the task of hairstyle transfer. In their work, they decompose an input latent code into a pair of latent codes representing structure and appearance. To transfer a given hairstyle, they blend between several images by taking specific regions of the structure latent codes and combining them with a target appearance. Finally, Chandran~\etal~\shortcite{chandran2021rendering} combine traditional and neural synthesis approaches by projecting high-quality skin maps into the latent space of StyleGAN, which is tasked with filling in regions that traditional methods struggle with --- such as the eyes, inner mouth, or hair. While all the aforementioned works showed incredible results and promise in real-world scenarios, they are limited in the domains they operate over. Some works have explored going beyond the facial domain and have explored applying StyleGAN for full-body synthesis in various applications such as virtual try-on and portrait reposing~\cite{Lewis2021TryOnGANBT,albahar2021pose}.

\vspace{0.25cm}
\section{Evaluation Metrics}~\label{sec:metrics}
While many aspects of GAN quality can be evaluated qualitatively, it is often desirable to assess the model quality more objectively. Evaluation metrics can be used to produce reliable, standardized benchmarks and to better gauge the advancement of the field. As we discuss below, this problem is not restricted to StyleGAN editing alone, but to the evaluation of most GANs and editing operations.

\paragraph*{GAN Evaluation}
The evaluation of generative models is straightforward when ground truth is at hand. For example, GAN inversion can be measured by various metrics assessing the distortion, such as pixel-wise distance using mean-squared error, perceptual similarity using LPIPS \cite{zhang2018unreasonable}, structural similarity using MS-SSIM \cite{wang2003multiscale}, or identity similarity~\cite{marriott2020assessment}, employed for facial images using a face recognition network \cite{deng2019arcface}. In the absence of such ground truth for the task of unconditional image synthesis, the evaluation of GAN quality remains an open challenge. Undoubtedly, the most popular metric is the Frechet Inception Distance (FID) \cite{heusel2017gans}. FID measures the similarity between two distributions using the Frechet Distance, where each distribution consists of visual features extracted by utilizing a pretrained recognition network ~\cite{szegedy2016rethinking}. Namely, given two sets of images, low FID indicates these sets share similar visual statistics. For the case of GANs, the target dataset is compared to the same number of random synthesized images, showing the similarity between these distributions.

Former to FID, Inception Score (IS) \cite{salimans2016improved} was introduced for the same purpose, measuring KL divergence over the same feature statistics. An additional approach has been suggested to measure the distance between real and generated images using the Sliced Wasserstein Distance (SWD) \cite{rabin2011wasserstein}, which computes the statistical similarity between local image patches extracted from the Laplacian pyramid of the images.
However, as FID is shown to be better correlated with the human perception of high-quality images, it has become the most widely used metric.

Despite its vast popularity, the FID metric does have drawbacks. As the extracted visual features are local, FID struggles to grasp a global structure. For facial images, which bear a simple structure, FID is still effective. Yet, images containing extremely unrealistic structures but high-quality textures, such as a cat with eight legs, can still achieve a good FID score undesirably. Another major concern is the employment of the popular truncation trick~\cite{marchesi2017megapixel,karras2019style}. Many works generate images using a latent truncation but measure FID without it, as it alters the distribution substantially and leads to a deterioration of FID values~\cite{katzir2022multilevel}. %

Sajjadi ~\cite{sajjadi2018assessing} proposed a solution to this exact problem by breaking down the GAN evaluation into recall and precision. High precision indicates high quality and realistic image generation, while high recall refers to generating a large amount of variation which is similar in diversity to the original data. 

\paragraph*{Editing Evaluation} 
For most practical cases, acquiring ground truth data and labeling to directly evaluate editing is infeasible or altogether impossible. As such, creative solutions have been proposed to tackle the problem of editing quality evaluation.
Contrary to disentanglement or GAN quality, the evaluation of StyleGAN's editing ability has not been widely studied. A few key aspects need to be analyzed for the evaluation of these editing procedures. Consider the example of adding a smile to a facial image. The most important aspect is the semantic meaning, namely, whether the editing successfully implants a smile. For binary editing, this could be easily performed using a classifier ~\cite{mokady2019masked, lample2017fader}, but in most cases, continuous editing is required. A regression model can be adopted for this case. However, for many attributes, these models are unavailable or require a vast amount of annotations to be trained. For example, recent works ~\cite{zhu2020improved, roich2021pivotal} used the Microsoft Face API ~\cite{azure} to measure face rotation but fail to measure the smile extent continuously. Furthermore, Zhu~\etal~\shortcite{zhu2020improved} demonstrate that the semantic editing magnitude when employing fixed editing is larger for the more native and editable regions of StyleGAN, and hypothesize that the magnitude could be utilized as an editability metric. 

Another key aspect is refraining from distorting the unedited parts of the image, usually referred to as preserving the original identity. For example, smile editing should not result in the appearance of glasses or a change in haircut. Some works ~\cite{Nitzan2020FaceID,richardson2020encoding,alaluf2021matter,tov2021designing, roich2021pivotal} focus on facial images, where identity preservation could be evaluated using facial recognition networks~\cite{deng2019arcface}. Since these networks are trained to be invariant to most attributes, adding a smile should not affect the output substantially. Therefore, an identity similarity can be measured by the cosine similarity of the facial identity representations. 
Nevertheless, as have been shown by Zhu~\etal~\shortcite{zhu2020improved}, the less editable latent spaces produce lower magnitude edits, leading to a bias in favor of these barely editable regions. Intuitively, the identity is better preserved better when the editing effect is reduced. Consequently, Roich~\etal~\shortcite{roich2021pivotal} suggest measuring the identity similarity while performing edits of the same magnitude, e.g. rotation to a predetermined angle. Such metrics have been shown to be more robust, with the identity similarity for the less editable $\mathcal{W+}$ space inversion being inferior compared to the native $\mathcal{W}$ space. 
Recent works \cite{yao2021latent, alaluf2021hyperstyle} have taken the above procedure a step further, plotting the measured identity similarity along a \textit{range} of editing magnitudes. This results in a continuous curve measuring identity preservation as a function of editing strength. In the context of videos, Tzaban~\etal~\shortcite{tzaban2022stitch} measured temporal coherence of edited videos, separating the evaluation of local (TL-ID) and global (TG-ID) identity consistency. Locally, they evaluate the identity similarity between pairs of adjacent frames. Globally, they measure similarity between all possible pairs, \ie{} not necessarily adjacent.

Still, these metrics are mostly limited to facial data, as it is challenging to procure identity recognition networks for other domains such as churches, cars, or cats. To this end, Abdal~\etal~\shortcite{abdal2020styleflow} focused on the setting of sequential editing and proposed to measure similarity between results obtained when applying the same semantic directions in a different order. Tov \etal~\shortcite{tov2021designing} suggested the latent editing consistency (LEC) metric to evaluate the editing quality realized by a given encoder $E$. Their method consists of performing latent editing followed by synthesis, encoding, and applying the reverse editing. Optimal editing is expected to result in minimal distortion as the editing procedure should only affect the desired attribute. 

One more concern is image quality. One of StyleGAN's key benefits is high visual quality, and editing methods should aim to preserve it. To this end, the common FID metric can be used over the edited images. However, editing might cause a significant bias between the edited and the real data, leading to inaccurate evaluation. If available, a classifier or regression model can be used to balance both image collections with respect to some attribute. 
A further approach, presented by Zhu~\etal~\shortcite{zhu2020improved}, is to evaluate the interpolation quality. They suggest that good editability should retain the high quality of StyleGAN even for the interpolated images, and utilize the FID metric for this purpose. 
Lastly, a number of works utilized a user study to evaluate editing quality~\cite{zhu2020improved, tov2021designing} through human judgment. Although this approach carries a profound understanding of the editing procedure, it consumes significant resources and is susceptible to unwanted manipulations. To this day, there is no widely acceptable assessment metric for latent manipulation quality.

\section{Discriminative Applications}
\label{sec:discriminative}

While the generative capabilities of GANs, and StyleGAN in particular, are indeed groundbreaking, one may ask what \textit{non-generative} tasks can potentially be tackled using GANs. In its most basic form, the GAN's capability to generate a large number of images, all essentially re-sampled from the same target distribution, can be used for data enrichment and augmentations for downstream training tasks. Indeed, many early works proposed using a GAN as an augmentation tool to generate more training data \cite{antoniou2017data,zhu2018emotion,tanaka2019data}, possibly also through the use of latent-space editing \cite{hochberg2021style}. 

Leveraging the GAN's editing capabilities, Chai \etal~\shortcite{chai2021ensembling} propose an ensembling method for image classification, by augmenting the input image at test-time. The input is projected into the pre-trained generator's latent space, and editing operations such as style mixing are applied to it, generating different views. The generated images are then fed into a classification network, and the final prediction of the model is based on an ensemble of the network predictions on all of the images. Unlike conventional ensembles in deep learning, where predictions of several models are combined to yield the final result, this method proposes using different views of the same image (while preserving its identity) and ensemble the classifier predictions on the images at test-time. 

\begin{figure}
\setlength{\tabcolsep}{1pt}
    \centering
    \includegraphics[width=\columnwidth]{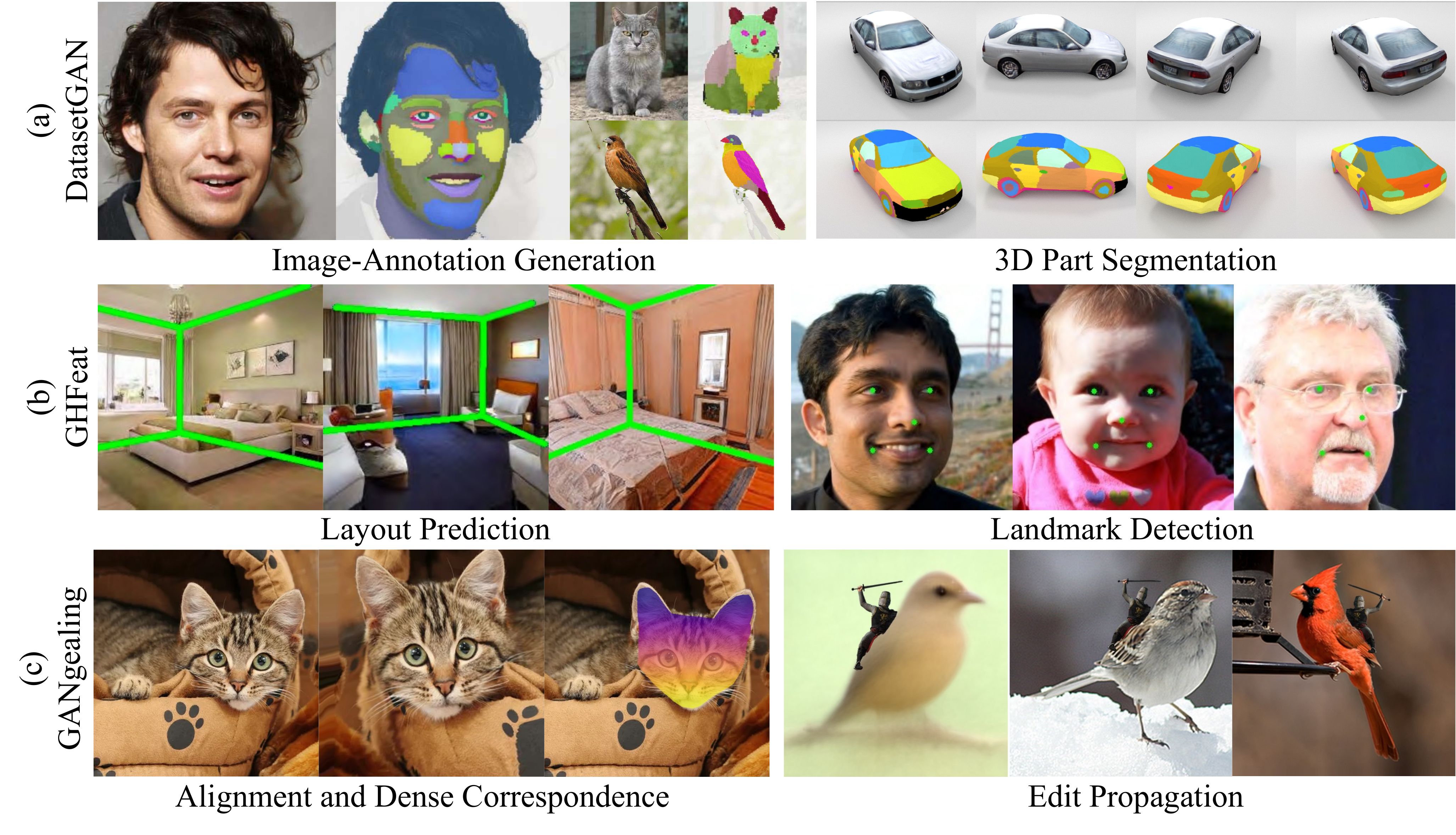}
    \vspace{-0.6cm}
    \caption{Examples of discriminative applications built around the StyleGAN generator. These include the ability to synthesize highly detailed segmentation masks (a)~\cite{zhang2021datasetgan}, regress facial landmarks or room layouts (b)~\cite{xu2021generative} and even learning alignments and dense correspondences which can be used to propagate edits between different images or video frames (c)~\cite{peebles2021gansupervised}.
    } 
    \vspace{-0.1cm}
    \label{fig:discriminative_applications}
\end{figure}

Aiming to leverage the semantic understanding of StyleGAN in new ways, Peebles~\etal~\shortcite{peebles2021gansupervised} present a novel framework to approach the task of dense visual alignment. Using sampled latent codes and their corresponding images, the authors jointly learn a latent representation used for latent-space editing, and a Spatial Transformer Network to align the generated images according to the edit, as illustrated in Figure~\ref{fig:discriminative_applications}c. 
Once both manipulations converge to a single viewpoint, the authors can employ the STN to align real images.
Abdal \etal~\shortcite{abdal2021labels4free} present an unsupervised segmentation method based on a pre-trained GAN. The authors recognize that nullifying some of the activations actually causes the GAN to erase the foreground object, producing an image with only a background. Hence, they form two networks, one that generates only the background for an image, and another which generates only the foreground, naturally leveraging the GAN's internal semantic understanding. They then use the two networks to train a segmentation mask generation network in an unsupervised manner. This notion, of extracting a segmentation map with the help of StyleGAN's structure, has been employed similarly by others
\cite{zhang2021datasetgan,li2021semantic,ling2021editgan,Lewis2021TryOnGANBT}.

However, StyleGAN's well-behaved latent space offers more opportunities. In a fine example of taking full leverage of the latent space, Xu~\etal~\shortcite{xu2021generative} show how to exploit pre-trained generative models for a wide variety of analysis and generation tasks. In essence, they propose an adversarial feature learning technique~\cite{donahue2016adversarial}; they extract meaningful feature maps through StyleGAN inversion and use them for a wide variety of tasks. The authors show that the channel-level modulations performed by a style code can be used as descriptive features for downstream tasks. Hence, by simply training an encoder, high-quality feature vectors can be produced in an unsupervised manner. The authors evaluate the quality of the features on different generative and discriminate downstream tasks, including image editing, image recognition, landmark detection, and more (see Figure~\ref{fig:discriminative_applications}b). Nitzan \etal~\shortcite{nitzan2021large} further observe that the linear nature of semantic directions in StyleGAN's latent space, \ie the same linearity exploited by the editing literature (see Section~\ref{sec:editing}), can be leveraged as a tool for few-shot regression. Their basic premise is that if the space is indeed linear, then given two labeled points along a disentangled axis, any interpolated point between them should produce an interpolation of their labels as well. In other words, they show that linear editing directions are not only global, in the sense that they cause a similar effect for all inputs, but also that the magnitude of these effects is linear in the size of the traversal step. Through this realization, they achieve state-of-the-art few-shot regression performance on various properties, such as yaw angle and age estimation for human faces. 

Continuing this line of thought, several papers leverage StyleGAN's intermediate representation to perform semantic segmentation. As previously discussed (see Section \ref{sec:endcoding}), a simple function, $f$, may be learned between the up-sampled concatenated per-layer features and a semantic segmentation of the image. As illustrated in Figure~\ref{fig:discriminative_applications}a, Zhang~\etal~\shortcite{zhang2021datasetgan} propose to use StyleGAN together with $f$ to generate a virtually infinite paired synthetic training set for semantic segmentation. Alternatively, Tritong~\etal~\shortcite{tritrong2021repurposing} directly use StyleGAN and $f$ for segmentation by first inverting a real image into latent space. 
In the context of local editing, Collins~\etal~\shortcite{collins2020editing} and Kafri~\etal~\shortcite{kafri2021stylefusion} perform a simple clustering procedure over StyleGAN's internal representations to obtain a semantic segmentation of an input image. This semantic map can then be used to perform local editing over an image, guided by a target reference image.
Lang~\etal~\shortcite{lang2021explaining} propose to not only exploit the emerging disentanglement properties of a pretrained StyleGAN, but to train a StyleGAN model for a specific disentangled axis. Through a clever training scheme, combining training StyleGAN along with a classifier for binary or multi-class recognition (e.g., a cat vs. dog classifier), they drive the latent space to capture classifier-specific attributes. As they demonstrate, through this joint training process, the linear editing directions that emerge from this model correspond to specific properties that the classifier searches for. For example, in the cat vs. dog case, the emerging editing directions include the shape of the eye, and the pointedness of the ears. This means that the image can be augmented to be more or less suitable for a specific label (e.g., a cat could be turned to be more dog-like), thus providing explainable examples of how the classifier makes its decision.  

As can be seen, StyleGAN's unsupervised arrangement of its latent space in disentangled directions is an exciting property that could potentially be leveraged for various applications. In the near future, it is likely that more works along these directions would be introduced, maybe establishing GANs as a method useful for many downstream tasks in machine learning in general, reaching beyond data augmentation or entertainment. Furthermore, it stands to reason that future generations of GANs may be designed with more consideration to discriminative tasks.
\section{Fine-Tuning the Generator}
\label{sec:fine-tuning}

\subsection{Data Reduction}
Training a StyleGAN model requires substantial amounts of data, confined to quite a small domain. This means that the amount of available data serves as a strong bottleneck for adapting StyleGAN to new domains.
An established way to address the lack of data is through augmentation. 
To stabilize training in limited data scenarios, several methods \cite{zhao2020diffAugment, Karras2020ada, zhao2020image} used differential augmentations during the training process. In contrast to classification tasks, generative augmentations pose a challenge --- if the discriminator observes sufficient augmented samples, the generator might produce such augmented results by itself. In many cases, the augmentations leakage to the generated images is highly undesirable, \eg when the augmentations contain rotations or unrealistic colorization changes. By monitoring overfitting indications during training and adaptively increasing augmentation strength, Karras~\etal~\shortcite{Karras2020ada} are able to introduce additional supervision to the network through augmentations, without allowing them to leak into the generated results. 
They achieved state-of-the-art results in low data domains, significantly reducing the number of training samples required for training.
Sinha~\etal~\shortcite{sinha2021negative} also suggest explicitly providing the discriminator with negative out-of-distribution samples to bias the generator away from unwanted samples. 
Another direction, proposed by Yang~\etal~\shortcite{yang2021insgen} is to empower the discriminator to better extract knowledge from the training set by providing it with an auxiliary task in the form of instance-discrimination via a contrastive learning objective. 
Kumari~\etal~\shortcite{kumari2021ensembling} 
propose to leverage the feature space of pre-trained vision models, trained for different vision tasks. By progressively selecting and employing the models as additional discriminators, they manage to improve synthesis quality in both limited-data and large-scale settings.

\begin{figure}
\setlength{\tabcolsep}{1pt}
    \centering
    \includegraphics[width=1.025\columnwidth]{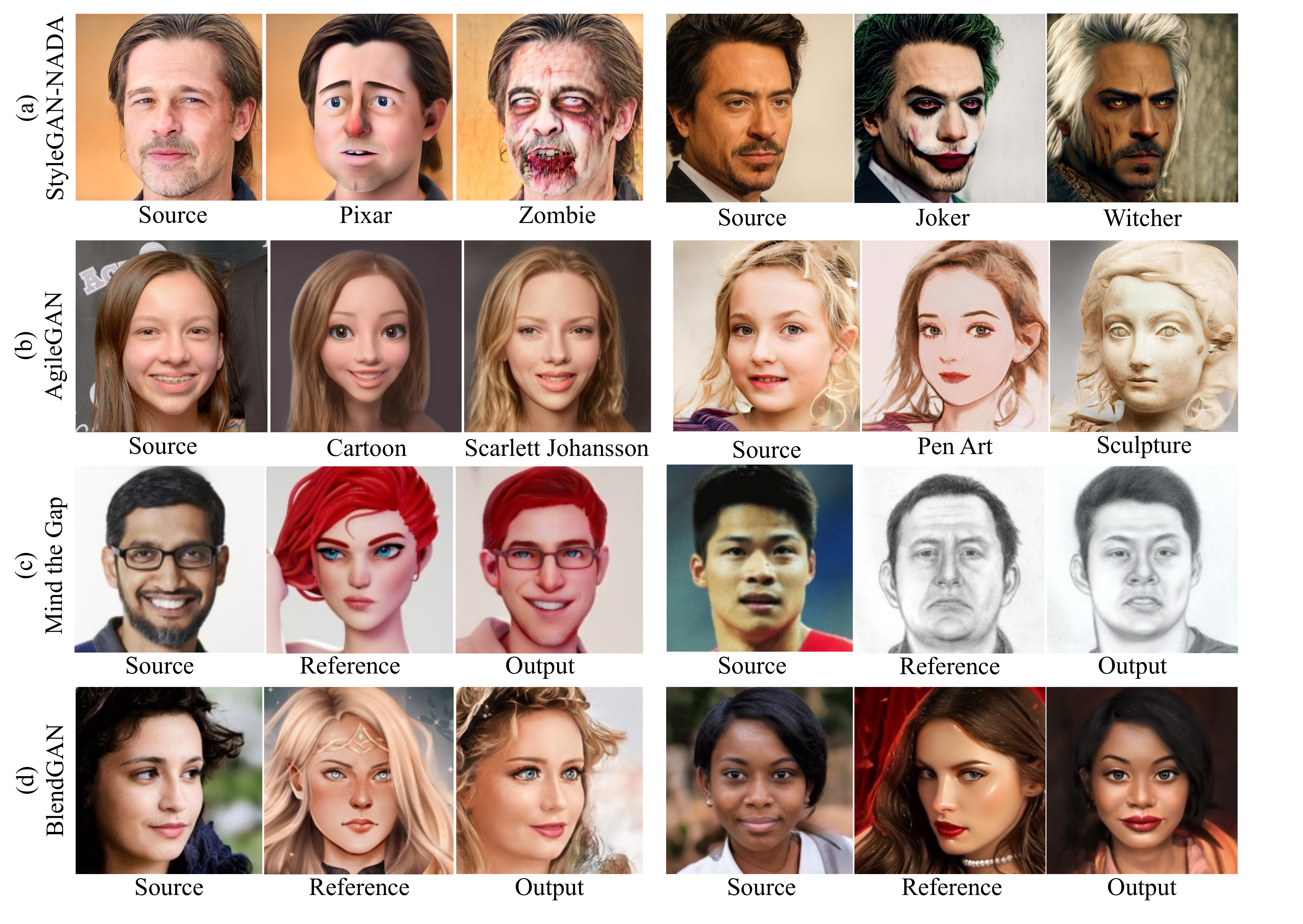}
    \vspace{-0.5cm}
    \caption{Many works have approached the task of transforming images from one domain (e.g., real faces) to other, semantically similar domains (e.g., cartoons). Typically, this has been done either through guidance via a textual description of the desired target domain (a)~\cite{gal2021stylegannada}, a short fine-tuning approach trained on a handful of images (b)~\cite{song2021agilegan}, or single-shot adaptation approach with style introduced via a desired reference image (c,d)~\cite{liu2021blendgan,zhu2021mind}.} 
    \label{fig:domain_adaptation}
\end{figure}

Another established and popular approach is \textbf{domain adaptation}, where different works seek to convert pre-trained StyleGAN models into other semantically similar domains using few data exemplars.
Aside from reducing the amount of data needed to train a model from scratch, Karras~\etal~\shortcite{Karras2020ada} fine-tuned a StyleGAN generator trained over the FFHQ dataset into the domain of MetFaces, a collection of images from the Metropolitan Museum of Art, using only $1,336$ training samples.
Following the above, Pinkney~\etal~\shortcite{pinkney2020resolution} leverage the
disentangled control over the coarse, medium, and fine semantic attributes StyleGAN offers and perform \textit{domain-mixing} for depicting the geometry of one domain with the textural appearance of another. 
The authors fine-tune a well-trained StyleGAN for human faces using only $300$ cartoon examples. They then propose blending the models, replacing high-resolution layers of the fine-tuned model with the pre-trained layers of the source domain. This yields a toonification effect, producing an output domain of cartoons, with the complex variety and structure of human faces. 
Several works make use of a similar form of fine-tuning.
Song~\etal~\shortcite{song2021agilegan} introduce several synthesis paths within the generator for different attributes, achieving high-quality portrait stylization.
Following the fine-tuning process, Jang~\etal~\shortcite{jang2021stylecarigan} leverage the semantic correspondence between the two models.
By feeding the same latent codes to both generators, an extensive paired training data set can be generated. 
This data is then used to train a translation network between the domains (in this case, a shape-exaggeration network operating over the caricatures domain).

Within the \textbf{few-shot} settings, several works perform domain adaptation based on as few as $10$ training exemplars. 
When the translation is done between semantically similar domains, the methods manage to preserve the semantic properties and diversity of the source domain.
Li~\etal~\shortcite{li2020few} maintain diversity by applying an elastic weight consolidation loss, regularizing weights change based on Fisher information, computed from a discriminator.
To facilitate few-shots adaptation, Ojha~\etal~\shortcite{ojha2021fewshotgan} propose explicitly maintaining the source domain structure, through a distance consistency loss between pairs of resulting images. In addition, a shared image and patch discriminator is applied to create patch-level adversarial similarity throughout the latent space together with image-level similarity around chosen anchors.

Taking domains with data shortage to the extreme, Gal~\etal~\shortcite{gal2021stylegannada} perform zero-shot domain adaptation, fine-tuning StyleGAN without providing any exemplar images. They propose describing the desired target domain in text and using a pre-trained linguistic-visual model~\cite{radford2021learning} to guide the adaptation. The GAN is trained such that the CLIP-space direction between the resulting images before and after the tuning process aligns with the CLIP-space direction between a pair of source and target description texts.
Zhu~\etal~\shortcite{zhu2021mind} perform single-shot domain adaptation by matching the reference image in the target domain with a corresponding synthesized image from the source domain, obtained through latent space optimization.
Using the reference pair, in addition to the CLIP-space loss defined in StyleGAN-NADA, they introduce a new objective, which is to maintain CLIP-space direction similarity between each reference image and the current training iteration's synthesized image. Using this method, the authors manage to achieve better adaptation of pose, lighting, and expression through the domain transfer process. Taking a different approach, Yang~\etal~\shortcite{yang2021oneshot}
freeze the generator weights and learn a linear transformation in $\mathcal{Z}$ space, using as little as a single reference image.
We refer the reader to Figure~\ref{fig:domain_adaptation} for sample domain adaptation results.

\subsection{Data-Aware Generator Tuning}

Arguably, the most promising direction for StyleGAN development is through data-aware model manipulation. A pre-trained StyleGAN model is phenomenal in local structure and disentanglement (see Section~\ref{sec:architecture}), but is relatively confined in generality. As such, it stands to reason that there could be significant benefit in adapting the model to include new, specific data points. Then, if the GAN's structure is maintained, these new points could be better processed for editing or discriminative applications. This concept has been suggested in the past already~\cite{bojanowski2018optimizing}. For example, in the context of GANs, Bau \etal~\shortcite{semantic2019bau} propose adapting the image prior learned by Progressive GAN~\cite{karras2017progressive} to image statistics of an individual image. Through minimal fine-tuning, the authors were able to faithfully reconstruct a given image, and present editing capabilities of quality unseen at the time, including synthesizing new objects seamlessly, removing unwanted objects, and changing object appearance. Similarly, Pan~\etal~\shortcite{pan2020exploiting} use BiGGAN~\cite{brock2018large} to capture high-level semantic image priors such as color, texture, spatial coherence, etc., for tasks such as colorization, inpainting, morphing, and category transfer. This is contrary to the traditional approach~\cite{ulyanov2018deep}, where only low-level priors are captured. Their method is based on GAN-inversion, but they overcome the difficulty of GAN-inversion methods to generate out-of-domain images by allowing the generator to be fine-tuned on the fly when searching for the latent source. They regularize the generator fine-tuning with feature matching loss from the discriminator, and use progressive fine-tuning (from shallow layers to deep).

In the context of StyleGAN, Roich~\etal~\shortcite{roich2021pivotal} take a similar approach. Given a real-world image that is similar, but not included in the domain of a pretrained StyleGAN (e.g., an image of a human face, with very untypical facial features, makeup, or hairstyle), they propose finding the closest latent code to the desired appearance (termed the `pivot'), and fine-tuning the generator so the exact appearance would be reconstructed with this code. In addition, they ensure the process does not impair the disentangled latent space through regularization. This simple approach produces significantly better reconstructions (see Fig.~\ref{fig:inv}), and allows employing off-the-shelf editing techniques with high quality, essentially bypassing the notorious distortion-editability trade-off (see Section~\ref{sec:encoding}). Such fine-tuning sessions are typically brief, lasting an order of a single minute.
As described in Section \ref{sec:inversion}, this generator tuning can alternatively be performed as a forward pass procedure using hypernetworks \cite{alaluf2021hyperstyle, dinh2021hyperinverter}.
Tzaban~\etal~\shortcite{tzaban2022stitch} further improve the tuning scheme to semantically edit a video while preserving temporal coherence. First, they observe that using an encoder for the initial inversion allows for a temporally-smooth edit after tuning the generator. Second, they propose to further tune the generator to better stitch the edited cropped face back to the original frame.

Bau~\etal~\shortcite{bau2020rewriting} perform a similar tuning operation, but where the awareness is to the task instead of the data. The authors propose changing semantic and physical properties (or \textit{rules}) of deep generative networks, relying on the concept of linear associative memory. While current methods for image editing allow users to manipulate single images, this method allows changing semantic rules and properties of the network, so that all images generated by the network have the desired property. This includes removing undesired patterns such as watermarks and adding objects such as human crowds or trees. Kwong~\etal~\shortcite{kwong2021unsupervised} outline a method for cross-domain editing by inverting images into a source domain and re-synthesizing them in a fine-tuned model using the same code. Cherepkov~\etal~\shortcite{cherepkov2021navigating} expand the range achievable by existing state-of-the-art generative models used for image editing and manipulation, such as StyleGAN2. While existing methods find interpretable directions in the model’s latent space and operate on latent codes, they find interpretable directions in the space of generator parameters and use them to manipulate images and expand the range of possible visual effects. They show that their discovered manipulations, such as changing car wheel size, cannot be achieved by manipulating the latent code. Finally, Liu~\etal~\shortcite{liu2022selfconditioned} demonstrate that brief fine-tuning sessions can be used to condition a model on labels derived from the latent-space itself, thereby ``baking'' editing directions into the GAN and improving treatment of rare data modalities, such as extreme poses or underrepresented ethnicities.

\vspace{-0.3cm}
\section{Conclusions}
\label{sec:conc}
StyleGAN has revolutionized the field of image synthesis, bringing with it consistent, high-quality results with exceptional photo-realism across multiple domains. More interestingly, through a combination of layer-wise style modulations and a novel mapping network, StyleGAN is capable of mapping out a smooth, semantic, and highly-disentangled latent space in an entirely unsupervised manner.
This enables latent-based editing, yielding effects such as photo-realistic and plausible alterations to age, hairstyles, or body poses, and even transformations into celebrities or magical beings. 

However, StyleGAN struggles with domains that do not exhibit strong structure. An in-depth look over the diverse set of works covered by this report will reveal that they all demonstrate their abilities on a rather limited collection of domains, and with no regard to the temporal axis. To address these limitations, there is yet much development that the generative field must undergo.

While many works focused on re-using a pre-trained generator for downstream tasks, a recent trend has shown that some of these domain-related limitations can be overcome if one adapts the generator to their specific needs. In essence, such approaches build upon the extensive knowledge that StyleGAN can glean from a rich source domain, and transfer it to new realms such as 3D rendering, paintings, or wildlife. 

Another noteworthy direction resides in extracting knowledge from StyleGAN for non-generative needs. StyleGAN has already been leveraged for regression, segmentation, and explainability, but there is doubtless more that could be learned from exploring its structured latent space. On the quest to self-supervision and learning representations that naturally disentangle and understand the elements comprising data distributions, StyleGAN is an important milestone.

\bibliographystyle{ACM-Reference-Format}
\bibliography{StyleGANStar}

\end{document}